\documentclass[accepted,mathfont=newtx]{uai2024} 
                        
\usepackage[american]{babel}

\usepackage{natbib} 
\usepackage{mathtools} 
\usepackage{booktabs} 
\usepackage{tikz} 

\usepackage[acronym,numberedsection,shortcuts,nonumberlist]{glossaries}
\setacronymstyle{long-sc-short}
\newacronym{BO}{BO}{Bayesian Optimization}
\newacronym{GP}{GP}{Gaussian Process}
\newacronym{ES}{ES}{Entropy Search}
\newacronym{EP}{EP}{Expectation Propagation}
\newacronym{EI}{EI}{Expected Improvement}
\newacronym{KG}{KG}{Knowledge Gradient}
\newacronym{RES}{RES}{Robust Entropy Search}
\newacronym{MES}{MES}{Max Value Entropy Search}
\newacronym{SSGP}{SSGP}{Sparse Spectrum Gaussian Process}
\newacronym{UCB}{UCB}{Upper Confidence Bounds}
\newacronym{JES}{JES}{Joint Entropy Search}

\usepackage{bm}

\newcommand*{\boldone}{\text{\usefont{U}{bbold}{m}{n}1}}

\usepackage{xcolor}

\usepackage{subcaption}

\usepackage{algorithm}
\usepackage{algpseudocode}

\DeclareMathOperator*{\argmax}{arg\,max}
\DeclareMathOperator*{\argmin}{arg\,min}

\newcommand{\sopt}{StableOpt}

\algrenewcommand\algorithmicrequire{\textbf{Input}}
\algrenewcommand\algorithmicensure{\textbf{Output}}

\title{Robust Entropy Search for Safe Efficient Bayesian Optimization}

\author[1]{\href{dorina.weichert@iais.fraunhofer.de}{Dorina Weichert}{}}
\author[2]{Alexander Kister}
\author[3]{Sebastian Houben}
\author[4,5]{Patrick Link}
\author[1]{Gunar Ernis}
\affil[1]{%
Fraunhofer Institute for Intelligent Analysis and Information
Systems IAIS\\ 
Sankt Augustin\\ 
Germany
}
\affil[2]{%
VP.1 eScience, Federal Institute for Materials Research and Testing BAM\\
Berlin\\
Germany
}
\affil[3]{%
University of Applied Sciences
Bonn-Rhein-Sieg\\
Sankt Augustin\\
Germany
}
\affil[4]{%
Fraunhofer Institute for Machine Tools and Forming Technology IWU\\
Chemnitz\\
Germany
}
\affil[5]{%
Institute of Mechatronic Engineering, TUD Dresden University of Technology\\
Dresden\\
Germany
}
  
\begin{document}
\maketitle
\begin{abstract}
  The practical use of \gls{BO} in engineering applications imposes special requirements: high sampling efficiency on the one hand and finding a robust solution on the other hand. 
  We address the case of adversarial robustness, where all parameters are controllable during the optimization process, but a subset of them is uncontrollable or even adversely perturbed at the time of application. 
  To this end, we develop an efficient information-based acquisition function that we call \gls{RES}.
  We empirically demonstrate its benefits in experiments on synthetic and real-life data. The results show that \gls{RES} reliably finds robust optima, outperforming state-of-the-art algorithms.
\end{abstract}

\section{Introduction}
\paragraph{Motivation}
\acrfull{BO} is a method for optimizing black-box functions that are costly to evaluate. It is used in various application domains, such as chemistry, robotics, or engineering \citep{shields2021bayesian,berkenkamp2023bayesian,lam2018advances}. 
The \gls{BO} framework consists of three ingredients: i) a Bayesian surrogate model of the unknown black-box function, traditionally a \gls{GP} regression model, ii) an acquisition function to specify the next evaluation point based on the surrogate model, and iii) the evaluation process of the black-box function.
Two fundamental properties that motivate the practical usage of \gls{BO} are a high sample efficiency (i.e., a fast convergence regarding the number of function evaluations) and robustness against noisy evaluations of the underlying black-box function \citep{garnett2023bayesian,Shahriari2016}.

The sample efficiency of \gls{BO} depends heavily on the choice of the acquisition function.
One class of acquisition functions are information-theoretic approaches, such as \gls{ES} \citep{Hennig2012}, Predictive Entropy Search \citep{Hernandez-Lobato2014}, \gls{MES} \citep{Wang17}, \gls{JES} \citep{Hvarfner2022}, and \(H_{l, A}\)-Entropy Search \citep{Neiswanger2022}. 
In all variations, the following evaluation point is chosen such that it maximizes the information gain about the (unknown) global optimum.
This line of reasoning is more sample-efficient than that of other acquisition functions, such as \gls{EI} \citep{Jones1998}, \gls{KG} \citep{frazier2008knowledge} or \gls{UCB}-based \citep{Srinivas2010} approaches but comes with higher computational cost \citep{garnett2023bayesian}.

While \gls{BO} is intrinsically robust against observation noise, as it is included into the surrogate model \citep{Shahriari2016, garnett2023bayesian}, engineering applications are often required to be adversarially robust. 
We face this requirement using a setting with two kinds of parameters: parameters \(\bm{x}\) that are controllable during the optimization process and at application time (\textit{controllable parameters}) and parameters \(\bm{\theta}\) that are controllable during the optimization process but externally affected at application time (\textit{uncontrollable parameters}).
A practical example of the latter set of parameters are environmental parameters, such as temperature, air pressure, or humidity, which are controllable in the lab but not at application time.
An adversarially robust solution solves the following objective function:
\begin{equation}
\label{eq:core_problem}
    \bm{x}^\star, \bm{\theta}^\star = \argmin_{\bm{x}} \argmax_{\bm{\theta}} f\left(\bm{x}, \bm{\theta}\right)~.
\end{equation}
It is an optimum of \(f\), which is minimal even under maximal negative perturbation by the uncontrollable parameter \(\bm{\theta}\).

We are the first to tackle this problem with a sample-efficient information-theoretic acquisition function, \acrfull{RES}.
Closest to our work are the approaches of \citet{Bogunovic}, who solve it by a \gls{UCB}-based approach, and of \citet{Frohlich2020}, who treat the related problem of mean-case robustness against input noise by an information-theoretic approach. 

\paragraph{Contributions} Our contributions can be summarized as follows: First, we formulate the conditions for an optimum being an adversarially robust one and integrate them into the heart of the acquisition function - the probability distribution over the function values conditioned on these requirements. Subsequently, we delineate a step-by-step approach for practically applying this intermediate result within an acquisition function. 
Lastly, we provide a rigorous empirical evaluation of our approach, utilizing synthetic data and real-world scenarios from robotics and engineering.

\section{Related Work}
Over the years, the traditional \gls{BO} setting for pure minimization (see, e.g., \citep{Shahriari2016, garnett2023bayesian} for overviews) was enhanced to match several robustness requirements.

Prevalent is the treatment of input perturbations, i.e., input uncertainty, via a mean measure \citep{Frohlich2020,beland2017,Nogueira2016,Iwazaki2021,Oliveira2019,Qing2022,Toscano-Palmerin2016,Toscano-Palmerin2022}: here, the objective is to minimize the expected value of an objective when the controllable parameters are perturbed, so to find \(\bm{x}^\star = \argmin_{\bm{x}} \mathbb{E}_{\bm{\theta} \propto p(\bm{\theta})} \left[f(\bm{x} + \bm{\theta}) \right]\). As a result, these approaches are more likely to find a broad instead of a narrow optimum.

In our work, we instead investigate a more conservative case: adversarially robust optimization that finds a worst-case optimal solution \(\bm{x}^\star = \argmin_{\bm{x}} \max_{\bm{\theta}}f(\bm{x}, \bm{\theta})\). \cite{Bogunovic} treated this case a special case of their groundbreaking \sopt{} algorithm that relies on the \gls{UCB} approach by \citet{Srinivas2010}. Superficially, adversarially robust optimization was also treated by \cite{Weichert2021} who adopt Thompson Sampling, \gls{ES} and \gls{KG} for discrete \(\bm{\theta}\). Recently, \cite{Christianson2023} introduced an adversarially robust version of \gls{EI}, dealing with a worst-case perturbation of the input, thus searching for the special case \(\bm{x}^\star = \argmin_{\bm{x}} \max_{\bm{\theta}} f(\bm{x} + \bm{\theta})\). In our approach, adding the input parameters is just one possible special case.

A further extension of the adversarially robust problem setting is distributionally robust optimization, where the goal is to find an optimum that is robust to a distributional shift within an uncertainty set \(U\) of an uncontrollable parameter: \(\bm{x}^\star = \argmin_{\bm{x}} \sup_{Q \in U} \mathbb{E}_{\bm{\theta} \propto Q} \left[f(\bm{x}, \bm{\theta})\right]\). The work of \citet{Kirschner2020} was the first approach to this problem utilizing \gls{UCB} until \citet{Husain2022,Tay2022,Yang2023} developed further approaches. Although these methods are related, they are not in the scope of our work. 

Only a few of the named approaches arise from the information-based acquisition functions. There are the method by \cite{Frohlich2020} to treat input perturbations and the one by \cite{Weichert2021} to treat adversarially robust entropy search for uncontrollable parameters from a discrete space.
Our contribution extends the existing research with an information-based adversarially robust acquisition function.

\begin{figure*}
    \centering
    \begin{subfigure}[b]{0.45\textwidth}
        \centering
        \includegraphics[width=1\linewidth]{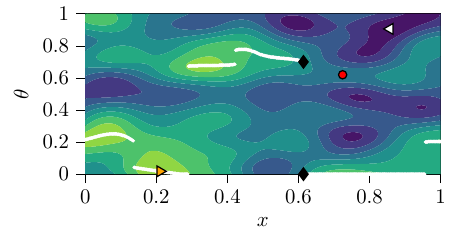} 
        \caption{objective function $f(\bm{x}, \bm{\theta})$.}
        \label{fig:1a}
    \end{subfigure}
    \hspace{0.05\textwidth}%
    \begin{subfigure}[b]{0.45\textwidth}
        \centering
        \includegraphics[width=1\linewidth]{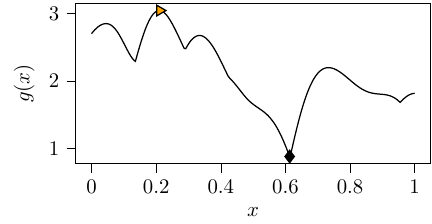}
        \caption{maximizing function $g(\bm{x})$, derived from $f(\bm{x}, \bm{\theta})$.}
        \label{fig:1b}
    \end{subfigure}
    \caption{Two-dimensional objective function $f(\bm{x}, \bm{\theta})$ and derived maximizing function $g(\bm{x}) = \max_{\bm{\theta}} f(\bm{x}, \bm{\theta})$. 
    In the given example, the location of the global robust optimum ($\blacklozenge$) is ambiguous. The optima are neither the global maximum (\textcolor{orange}{$\blacktriangleright$}), the global minimum ($\lhd$) nor the smallest local min max point (\textcolor{red}{$\bullet$}). The values of the argmax function \(\bm{h}(\bm{x})\) are rendered as a white line in figure~\ref{fig:1a}. The function values at these points define the maximizing function \(g(\bm{x})\), given in figure~\ref{fig:1b}.}
    \label{fig:1}
\end{figure*}

\section{Background}
Before we delve deeper into the derivation of the acquisition function, we would like to revisit \gls{GP}s and briefly explain some basic properties of the adversarially robust optimum.
\subsection{Gaussian Process Regression}
\gls{GP} regression is a non-parametric method to model an unknown function \(f(\bm{z}): \mathcal{Z} \mapsto \mathbb{R}\) by a distribution over functions.
The \gls{GP} prior is defined such that any subset of function values is normally distributed with mean \(\mu_0(\bm{z})\) and covariance \(k(\bm{z}, \bm{z}')\) for any \(\bm{z}, \bm{z}' \in \mathcal{Z}\) (w.l.o.g. we assume \(\mu_0(\bm{z}) = 0\) \citep{Rasmussen2006}). 
Conditioning the prior on actual data \(D_t = \lbrace (\bm{z}_1, y_1), \dots, (\bm{z}_t, y_t)\rbrace\), where \(\bm{y} = f(\bm{z}) + \epsilon\), \(\epsilon \sim \mathcal{N}(0, \sigma_n)\), the predictive posterior distribution \(p(f) \sim GP(m_t, v_t \vert D_t)\) is given by 
\begin{equation}
\label{eq:GP_definition}
    \begin{split}
        m_t(\bm{z} \vert D_t) &= \bm{k}(\bm{z})^T \bm{K}^{-1} \bm{y} \\
        v_t(\bm{z} \vert D_t) &= k(\bm{z}, \bm{z}) - \bm{k}(\bm{z})^T \bm{K}^{-1} \bm{k}(\bm{z})~,
    \end{split}
\end{equation}
with \(\left[\bm{k}(\bm{z})\right]_i = k(\bm{z}, \bm{z}_i)\), \(\bm{K}_{i,j} = k(\bm{z}_i, \bm{z}_j) + \delta_{ij}\sigma^2_n\), where \(\delta_{ij}\) is the Kronecker delta, and \(\left[\bm{y}\right]_i = y_i\). 

\gls{GP}s are common surrogate models in \gls{BO}. 
Since we consider a set of controllable parameters \(\bm{x} \in \mathcal{X} = \mathbb{R}^{d_c}\) and a set of uncontrollable parameters \(\bm{\theta} \in \Theta = \mathbb{R}^{d_u}\), \(\bm{z}\) in the previous definitions is replaced by the concatenation of \(\bm{x}\) and \(\bm{\theta}\), in our case \(\mathcal{Z} = \mathcal{X} \times \Theta\).

\subsection{Properties of the Robust Optimum}
\label{sec:robust_properties}
The robust optimum \((\bm{x}^\star, \bm{\theta}^\star)\) has to fulfill two nested conditions:
\begin{itemize}
    \item[(a)] \textbf{Its function value is maximal in the direction of the uncontrollable parameters \(\bm{\theta}\),} generating a maximizing function \(g(\bm{x}) = \max_{\bm{\theta}} f(\bm{x}, \bm{\theta})\) and an argmax function \(\bm{h}(\bm{x}) = \argmax_{\bm{\theta}} f(\bm{x}, \bm{\theta})\).
    \item[(b)] \textbf{The optimum minimizes the maximizing function \(g(\bm{x})\)}. In consequence, the robust minimum is generally neither the global maximum nor minimum, but there generally exist function values of \(f\) that are smaller and function values of \(f\) that are larger than the robust optimum.   
\end{itemize}

The difference between the optima is visualized as an example in figure~\ref{fig:1}.
Besides of the global robust optimum ($\blacklozenge$), we show the global maximum (\textcolor{orange}{$\blacktriangleright$}), the global minimum ($\lhd$) and the smallest local min max point (\textcolor{red}{$\bullet$}) (Nash equilibrium). Neither of the latter optima corresponds to the robust optimum that is sought.

\section{Robust Entropy Search}
We propose the \gls{RES} acquisition function that considers the properties of the robust optimum by involving the noiseless robust optimal value \(f^\star =f(\bm{x}^\star, \bm{\theta}^\star)\), the argmax function \(\bm{h}(\bm{x})\) and its corresponding function values \(g(\bm{x})\). Throughout the section we call these three quantities, \((\bm{h}, g, f^\star)_f\) that all depend on \(f\), \textit{robustness characteristics}.

\subsection{Methodical Idea}
Like other information-based acquisition functions, see, e.g., \gls{MES}~\citep{Wang17} or \gls{JES}~\citep{Hvarfner2022}, \gls{RES} deduces the optimum by means of mutual information \(I\) between the value $y(\bm{z}) = f(\bm{z}) + \varepsilon$ at the proposed location $\bm{z}$ and some property of the optimum, in our case, the robustness characteristics  \((\bm{h}, g, f^\star)_f\).
\gls{RES} follows
\begin{equation}
\label{eq:Mutual_information}
    \begin{split}
        &\alpha_{RES}(\bm{z}) = I\left(\left(\bm{z},y \right), \left(\bm{h}, g, f^\star\right)_f \vert D_t\right) \\
        &= H\left[ p\left(y\left(\bm{z}\right) \vert D_t\right) \right] \\ 
        &~~~~~~~~~~~~ - \mathbb{E}_{\left(\bm{h}, g, f^\star\right)_{f}} 
        \left[ H\left[p\left(y\left(\bm{z}\right) \vert \left(\bm{h}, g, f^\star\right)_{f} , D_t\right)\right]\right] \\
        &\approx H\left[ p\left(y\left(\bm{z}\right) \vert D_t\right) \right]  \\ 
        &~~~~~~~~~~~~ - \frac{1}{C}\sum_{
        f_c \in \mathcal{F}_c} 
        H\left[p\left(y\left(\bm{z}\right) \vert \left(\bm{h}_c, g_c, f_c^\star\right)_{f_c}, D_t\right)\right]~,
    \end{split}    
\end{equation}
where \(\mathcal{F}_c\) is a set of \(C\) functions sampled from the actual \gls{GP} posterior \(GP(m_t, v_t \vert D_t)\) for the purpose of approximation. For each individual sample \(f_c \in \mathcal{F}_c\), we find the corresponding robustness characteristics \((\bm{h}_c, g_c, f^\star_c)_{f_c}\). 
As these quantities follow a joint distribution, only one expectation is taken.

As we not only involve \(f^\star\) but also the argmax function \(\bm{h}(\bm{x})\) and the maximizing function \(g(\bm{x})\), the acquisition function proposes points that are likely to reduce the uncertainty about all robustness characteristics simultaneously.

The approximation of the conditional distribution $p(y(\bm{z}) \vert \left(\bm{h}_c, g_c, f^\star_c\right)_{f_c},  D_t)$ lies at the center of the acquisition function. In a first step, we simplify it by approximating noisy \(y\) with \(f\), since the observation noise is additive and can be added later when computing the entropy. Secondly, we implement the conditions formulated in section~\ref{sec:robust_properties} into the conditional distribution. Therefore, we use indicator functions denoted by \(\boldone_{\lbrace \cdot\rbrace}\):
\begin{equation}
\label{eq:conditioned_probability}
    \begin{split}
        &p\left(f \vert \left(\bm{h}_c, g_c, f^\star_c\right)_{f_c},  D_t,\right) \\
        &\propto\int\mathrm{d}f \, p(f\vert D_t) \cdot \boldone_{\lbrace f\left(\bm{x}, \bm{\theta}\right) \leq g_c\left(\bm{x}\right) \rbrace} \\
        &~~~~~~~~~~\cdot \boldone_{\lbrace f^\star_c \leq f\left(\bm{x}, \bm{h}_c(\bm{x})\right) \leq g_c\left(\bm{x}\right)\rbrace} \\
    \end{split}
\end{equation}
The first indicator function implements the requirement of the optimum to be the maximum over the uncontrollable parameters \(\bm{\theta}\), referring to condition (a). By the second indicator function, we aim to find the minimum of these maxima by using the sampled optimum \(f_c^\star\) as a lower bound on the distribution of maximum function values, implementing condition (b).
Equation~(\ref{eq:conditioned_probability}) is already a simplification and approximation of the actual target in equation~(\ref{eq:Mutual_information}): Instead of conditioning on the whole extreme functions \(\bm{h}\) and \(g\), we only condition on the values of these functions at \((\bm{x}, \bm{\theta})\). 

\subsection{Implementation}
Our approach relies on the efficient treatment of samples from a \gls{GP} and on the efficient calculation of the posterior predictive distribution, conditioned on the robustness characteristics. We summarize all necessary implementation steps in the following.
\subsubsection{Efficient Treatment of Function Samples}
To efficiently sample from the actual \gls{GP}, we make use of the \gls{SSGP} approximation by \citet{lazaro-gredilla10a}, which offers the opportunity to draw \gls{GP} samples that have a closed analytical expression.
This is beneficial for our approach, as we have to find the robustness characteristics numerically. Samples formed by this \gls{GP} approximation are effectively optimized using gradient descent methods as derivatives are also available.

The function samples are of the form \(f_c(\bm{z}) = \bm{a}^T \bm{\phi} (\bm{z})\), with weight vector \(\bm{a}\) and a vector of feature functions \(\bm{\phi}(\bm{z}) \in \mathbb{R}^F\), where \(F\) is the number of feature functions. 
The elements \(i\) of the feature vector \(\bm{\phi}\) are given by \(\phi_{i}\left(\bm{z} \right) = \cos \left(\bm{w}_i^T \bm{z} + b_i\right)\) with \(b_i \sim U(0, 2\pi)\) and \(\bm{w}_i \sim p(\bm{w}) \propto s(\bm{w})\) where \(s(\bm{w})\) is the Fourier dual of the covariance function \(k\).
The elements of the weight vector \(\bm{a}\) follow a normal distribution \(\mathcal{N}\left(\bm{A}^{-1}\bm{\Phi}^T\bm{y}, \sigma^2_{n} \bm{A}^{-1}\right)\), with \(\bm{A} = \bm{\Phi}^T\bm{\Phi} + \sigma^2_{n} I\), \(\bm{\Phi}^T\) being the matrix composed from the feature function evaluated at the input data \(\bm{\Phi}^T = \left[\bm{\phi}(\bm{z}_1), \dots, \bm{\phi}(\bm{z}_t) \right]\), and \(\bm{y}\) being the corresponding observed function values. 
The sampling of functions therefore takes place in two steps: First, we draw frequencies \(\bm{w}_i\) and phases \(b_i\) to generate an unbiased approximation of the covariance function \citep{Rahimi2007}. We then draw as many weight vectors \(\bm{a}\) as function samples are needed from the resulting normal distribution. The resulting function samples can be evaluated cost-effectively by simple matrix-vector multiplication.
For more details, see, e.g. the work of \citet{lazaro-gredilla10a} or \citet{Hernandez-Lobato2014}.

To find the argmax function \(\bm{h}_c(\bm{x})\) and maximum value \(g_c(\bm{x})\), a standard numerical solver, e.g. a gradient descent method, is called on-the-fly. 
To find the robust optimum, we implement a nested numerical solver that calculates the actual maximum over the uncontrollable parameters at every minimization step over the controllable ones.

\subsubsection{Calculating the Conditioned Posterior Probability Distribution}
\begin{figure*}
    \begin{subfigure}[t]{0.45\textwidth}
        \centering
        \includegraphics[width=1\linewidth]{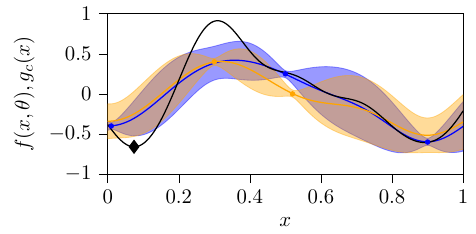} 
        \caption{predictive distribution and sample before conditioning}
        \label{fig:3a}
    \end{subfigure} \hspace{0.075\textwidth}%
    \begin{subfigure}[t]{0.45\textwidth}
        \centering
        \includegraphics[width=1\linewidth]{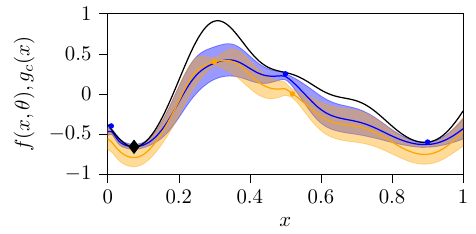} 
        \caption{predictive distribution after conditioning on the sample \(g_c(x)\) and the robust sample optimum $f^\star_c$}
        \label{fig:3b}
    \end{subfigure}
    \caption{Predictive distributions (mean and one standard deviation) before and after conditioning for a single uncontrollable parameter with two values \(\mathbb{\theta}_1\) (blue) and \(\mathbb{\theta}_2\) (orange). In this case, \(\bm{h}_c(\bm{x}) = \mathbb{\theta}_1~\forall~\bm{x}\) (blue) with the max function $f_c(\bm{x}, \theta_1) = g_c(\bm{x})$ (black). While the predictive distribution \(f(\bm{x}, \mathbb{\theta}_2)\) is only upper bounded by the sample \(g_c(\bm{x})\), \(f(\bm{x}, \mathbb{\theta}_1)\) is upper bounded by the sample  \(g_c(\bm{x})\) and lower bounded by the optimum \(f_c^\star\) (\(\blacklozenge\)).}
    \label{fig:3}
\end{figure*}
A key ingredient for \gls{RES} is the calculation of the conditional probability \(p(f \vert \left(\bm{h}_c, g_c, f^\star_c\right)_{f_c}, D_t)\) in equation (\ref{eq:conditioned_probability}). 
As directly working on the function space is complex, we take a three-step approach to approximate the conditioned posterior probability distribution, inspired by the ideas of \citet{Frohlich2020} and \citet{hoffman2015output}. 
Our final approximation is normal-distributed, and we can leverage the fact that the entropy of a normal distribution is given analytically for calculating the acquisition function.

\paragraph{Step 1: Conditioning the \gls{GP} at the training data points.}

Instead of taking into account the whole \gls{GP} on \({\mathcal{X} \times \Theta}\), we consider it only on a discrete subset of points from \(\mathcal{X} \times \Theta\): the already evaluated training data points \(D_t\).
We enforce equation~(\ref{eq:conditioned_probability}) to be true for all \(\bm{z}_i = (\bm{x}_i, \bm{\theta}_i) \in D_t\). Therefore, after calculating the maximizing uncontrollable parameters \(\bm{h}_c(\bm{x}_i)\) and their corresponding function values \(g_c(\bm{x}_i) = f_c(\bm{x}_i, \bm{h}_c(\bm{x}_i))\), 
we condition \(\bm{f} = \left[f(\bm{z}_1),\dots,f(\bm{z}_t),f(\bm{x}_1, \bm{h}_c(\bm{x}_1)),\dots,f(\bm{x}_t, \bm{h}_c(\bm{x}_t)) \right]^T\) on the robustness characteristics by \gls{EP} \citep{Minka2001}:
\begin{equation}
\label{eq:conditioned_train_data}
    \begin{split}
        &p\left(\bm{f} \vert \left(\bm{h}_c(\bm{x}), g_c(\bm{x}), f^\star_c\right)_{f_c}, D_t\right) \\
        &\propto p\left(\bm{f} \vert D_t\right) \prod_{i=1}^{t} \boldone_{\lbrace f\left(\bm{x}_i, \bm{\theta}_i\right) \leq g_c(\bm{x}_i) \rbrace} \\
        &~~~~~~~~~~~~~~~~~~~~~~~~~\cdot 
        \boldone_{\lbrace f^\star_c \leq f\left(\bm{x}_i, \bm{h}_c\left(\bm{x}_i\right)\right) \leq g_c(\bm{x}_i)  \rbrace}\\
        &\stackrel{\text{(EP)}}{\approx:} \mathcal{N}(\bm{\mu}_1, \bm{\Sigma}_1) ~.    
    \end{split}
\end{equation}
We approximate by \gls{EP}, because problems of the form of equation~(\ref{eq:conditioned_train_data}) can only be solved analytically for lower dimensions \citep{Rosenbaum1961,ang2002}.
\gls{EP} has been shown to efficiently approximate the required measures in a reasonable computation time \citep{Frohlich2020,Gessner2020,Hennig2012}. 
We reuse the implementation for linearly constrained Gaussians by~\cite{Frohlich2020}, building on the work of~\cite{Herbrich} and reformulate the indicator functions to lower bounds \(\bm{l}_b = \left[\bm{0}^{(1 \times t)},f_c^{\star{(1 \times t)}}\right]^T\) and upper bounds \(\bm{u}_b = \left[g_c(\bm{x}_1),\dots,g_c(\bm{x}_t),g_c(\bm{x}_1),\dots,g_c(\bm{x}_t)\right]^T\) to find the approximation \(\mathcal{N}(\bm{\mu}_1, \bm{\Sigma}_1)\).
            
\paragraph{Step 2: Creating a predictive distribution for a new location \(\bm{z}\).}

We obtain a predictive distribution by marginalizing over the function values \(\bm{f}\), using \gls{GP} arithmetic. Already looking ahead to step three, we predict at \({\hat{\bm{z}}=\left[(\bm{x}, \bm{\theta}),(\bm{x},\bm{h}_c(\bm{x}))\right]^T}\), receiving predictions ${p(f(\bm{\hat{\bm{z}}}) \vert D_t, \bm{f}) = \mathcal{N}(\bm{\mu}_f, \bm{\Sigma}_f)}$. We find
\begin{equation}
    \begin{split}
        p_0(f(\hat{\bm{z}}) &\vert \left(\bm{h}_c(\bm{x}), g_c(\bm{x}), f^\star_c\right)_{f_c},  D_t) \\
        &= \int p(\bm{f} \vert \left(\bm{h}_c(\bm{x}), g_c(\bm{x}), f^\star_c\right)_{f_c}, D_t) \\ 
        &~~~~~~~~~~~~~~~~~~~~~~~~~~~~~~~~~~~\cdot p(f(\bm{\hat{\bm{z}}}) \vert D_t, \bm{f}) \mathrm{d} \bm{f} \\
        &\approx \int \mathcal{N}(\bm{\mu}_1, \bm{\Sigma}_1) \cdot \mathcal{N}(\bm{\mu}_f, \bm{\Sigma}_f) \, \mathrm{d}\bm{f} \\
        &\approx \mathcal{N}(f(\hat{\bm{z}}) \vert m_0(\hat{\bm{z}}), v_0(\hat{\bm{z}}))~.
    \end{split} 
    \label{eq:prediction}
\end{equation}%
The predictive distribution again follows a normal distribution.
    
\paragraph{Step 3: Conditioning the predictions.}
    
As we only required the robustness conditions to be true for the training set \(D_t\) in step~1, we now apply them to the predictions:
\begin{equation}
    \begin{split}
        &p(f(\hat{\bm{z}}) \vert \left(\bm{h}_c(\bm{x}), g_c(\bm{x}), f_c^\star\right)_{f_c}, D_t) \\
        &= \mathcal{N}(f(\hat{\bm{z}}) \vert m_0(\hat{\bm{z}}), v_0(\hat{\bm{z}})) \cdot 
        \boldone_{\lbrace f\left(\bm{x}, \bm{\theta}\right) \leq g_c(\bm{x}) \rbrace} \\
        &~~~~~~~~~~~~~~~~~~~~~~~~~~~~~~~~~~~~~~~~\cdot 
        \boldone_{\lbrace f_c^\star \leq f\left(\bm{x}, \bm{h}_c\left(\bm{x}\right)\right) \leq g_c(\bm{x}) \rbrace} \\
        &\approx \mathcal{N}(f(\hat{\bm{z}}) \vert \hat{m}_q(\hat{\bm{z}}), \hat{v}_q(\hat{\bm{z}}))
    \end{split}
\end{equation}
For a single \(\bm{z}\), we find a bivariate doubly truncated Gaussian with bounds as in step 1, for which the matching first and second moments are known analytically \citep{ang2002} and given in appendix~\ref{sec:doubly_truncated}. From the matching moments, we extract the ones corresponding to the original \(\bm{z}\) by indexing: \(m_q(\bm{z}) = \hat{m}_{q(0)}\), \(v_q(\bm{z}) = \hat{v}_{q(0, 0)}\).    

In figure~\ref{fig:3}, we show the effect of conditioning for a problem with one discrete uncontrollable parameter \(\bm{\theta} = \lbrace\theta_1, \theta_2 \rbrace\) with two possible values. The worst-case function sample \(g_c(\bm{x})\) originates from the blue uncontrollable parameter value \(\mathbb{\theta}_1\). 
The resulting posterior predictive distribution is changed as follows: on the one hand, all function values are upper-bounded by the maximizing function sample; on the other hand, the function values of \(f(\bm{x}, \mathbb{\theta}_1)\) are additionally lower-bounded by the sampled optimal value \(f_c^\star\).

\subsubsection{Final formulation of the RES acquisition function.}
Given the final approximation \((m_q(\bm{z}), v_q(\bm{z}))\),
we formulate the \gls{RES} acquisition function as
\begin{equation}
    \begin{split}
        \alpha_{\text{RES}}(\bm{z}) &= \frac{1}{2} \log \left(\left(v_t \left(\bm{z} \right)\vert D_t\right)  + \sigma^2_n \right) \\
        &- \frac{1}{2C}  \\
        \cdot \sum_{f_c \in \mathcal{F}_c } &\log \left(\left( v_q\left(\bm{z}\right)\vert \left(\bm{h}_c(\bm{x}), g_c(\bm{x}), f^\star_c\right)_{f_c}, D_t\right) + \sigma^2_n \right)~.     
    \end{split}
\end{equation}
We summarize all necessary optimization steps in algorithms~\ref{alg:robust_optimization} and \ref{alg:RVES}.
In each iteration \(t\), an \gls{SSGP} approximation of the actual \gls{GP} is calculated, and \(C\) function samples are drawn. The robust optima \(f^\star_c\) are calculated for these samples. Then, the \gls{GP} is conditioned on the resulting robustness characteristics at the actual training data points. This step is only performed once when optimizing the acquisition function.
Afterward, creation of the predictive distribution and conditioning at the new point \(\bm{z}\) is performed individually for each \(\bm{z}\) that is called during optimization of the acquisition function. Finally, we return the robust optimum of the actual model's predictive mean \(m_t\).

In appendix~\ref{sec:theoretical_runtime}, we provide results on the time complexity of our algorithm for different combinations of discrete and continuous variables. Overall, the runtime is governed by the calculation of equation~(\ref{eq:prediction}), with effects from calculating the argmax function \(\bm{h}_c(\bm{x})\) and the \gls{GP} prediction to obtain $\mathcal{N}(\bm{\mu}_f, \bm{\Sigma}_f)$.

\begin{algorithm}
\caption{Robust \gls{BO} with \gls{RES} acquisition function.}\label{alg:robust_optimization}
\begin{algorithmic}[1]
\Require{maximum number of iterations $T$, space of controllable parameters \(\mathcal{X}\), space of uncontrollable parameters \(\Theta\), number of samples $C$, size of initial design $M$}
\Ensure{robust optimum $\bm{z}^\star = (\bm{x}^\star, \bm{\theta}^\star)$}
\State $D_{M} \gets \lbrace \bm{z}_i, \bm{y}_i \rbrace_{i=1}^M$
\For{$t=M,\dots,M+T-1$}
    \State $GP(m_t(\bm{z}), v_t(\bm{z})) \gets \Call{FitGP}{D_t}$
    \State $\mathcal{F}_c \gets \Call{SampleGP}{GP, C}$ \Comment{Create \gls{SSGP}, sample}
    \State $\mathcal{F}_c^\star \gets \emptyset$ 
    \For{$c=1,\dots,C$}       
        \State $\mathcal{F}_c^\star \gets \mathcal{F}_c^\star \cup f_c^\star = \min_{\bm{x} \in \mathcal{X}} \max_{\bm{\theta} \in \Theta} f_c\left(\bm{x}, \bm{\theta} \right)$                
    \EndFor
    \State $\bm{z}_{t+1} \gets \argmax_{\bm{z} \in \mathcal{X} \times \Theta}  \alpha_{RES}(\bm{z}, GP, \mathcal{F}_c, \mathcal{F}_c^\star)$
    \State $\bm{y}_{t+1} = f(\bm{z}_{t+1}) + \epsilon, D_{t+1} \gets D_t \cup \lbrace \bm{z}_{t+1}, \bm{y}_{t+1}\rbrace$
\EndFor
\State \textbf{return} $(\bm{x}^\star, \bm{\theta}^\star) \gets \argmin_{\bm{x} \in \mathcal{X}} \argmax_{\bm{\theta} \in \Theta}{m_t(\bm{x}, \bm{\theta})}$
\end{algorithmic}
\end{algorithm}

\begin{algorithm}
\caption{The RES acquisition function.}\label{alg:RVES}
\begin{algorithmic}[1]
\Require{evaluation point $\bm{z}$, \gls{GP} $GP$, function samples $\mathcal{F}_c$, robust optima $\mathcal{F}_c^\star$}
\Ensure{value of \gls{RES} acquisition function}
\State $H \gets 0$
\For{$c \in \lbrace1,\dots,C\rbrace$}
\If{$\Call{IsNotInitialized}{\alpha_{RES}}$} 
\State $\bm{\mu}_1, \bm{\Sigma}_1 \gets \Call{ApproximateEP}{GP, f_c, f^\star_c}$\\
\Comment{sec. 3.2.2., step 1}
\EndIf
\State $\bm{h}_c(\bm{x}) \gets \argmax_{\bm{\theta} \in \Theta} f_c(\bm{x}, \bm{\theta})$
\State $g_c(\bm{x}) \gets f_c(\bm{x}, \bm{h}_c(\bm{x}))$
\State $v_q(\bm{z}) \gets$ \\
$\Call{ConditionPosteriorVariance}{\bm{\mu}_1, \bm{\Sigma}_1, \bm{h}_c, g_c}$\\
\Comment{sec. 3.2.2, steps 2 \& 3}
\State $H \gets H + \log(v_q(\bm{z}) + \sigma_n^2)$ 
\EndFor
\State \textbf{return} $\alpha_{RES} \gets \frac{1}{2} \log (v_t(\bm{z}) + \sigma_n^2) - \frac{1}{2C} H$
\end{algorithmic}
\end{algorithm}

\section{Experiments}
We conduct three types of experiments: in a preliminary test, we estimate the general performance of the acquisition function and its dependency on the number of necessary function samples \(C\) in a within-model comparison.
Secondly, we compare our algorithm with state-of-the-art benchmarks on synthetic problems. 
Finally, two real-life problems are treated: the calibration of parameters of a numerical simulation, arising in an engineering task, and robust robot pushing, an experiment formulated by \cite{Bogunovic}.

We compare our approach, \gls{RES}, to \sopt{} \citep{Bogunovic} with different exploration constants \(\sqrt{\beta}\), and with the non-robust acquisition functions \gls{MES} \citep{Wang17}, \gls{UCB} \citep{Srinivas2010}, \gls{KG} \citep{frazier2008knowledge}, and with standard \gls{EI} \citep{Jones1998}. 
In \gls{RES}, we set the number of features \(F = 500\) for the \gls{SSGP}.
For \gls{MES}, we choose a value of a number of 100 sampled minima, and the exploration parameter \(\sqrt{\beta}\) in \gls{UCB} was set to a value of 2. 
For \gls{KG}, which was originally designed for discrete spaces, we discretize the continuous space of dimensionality \(d_{\text{conti.}}\) by a random grid of size \(50^{d_{\text{conti.}}}\) drawn from a uniform distribution in each iteration and use a number of 32 function samples.
For StableOpt, based on the experiments in the original publication, we apply constant exploration constants from \(\sqrt{\beta} \in \lbrace1,2,4\rbrace\).

For measuring performance, we use algorithm- and problem-specific metrics.
As \gls{RES} evaluates at a location that raises the knowledge about the optimum and not at a potential optimum location, the optimum location is calculated at every iteration as the robust optimum of the actual model mean \(\bm{z}_t^\star = (\bm{x}^\star, \bm{\theta}^\star) = \argmin_{\bm{x} \in \mathcal{X}} \argmax_{\bm{\theta} \in \Theta} m_t(\bm{x}, \bm{\theta})\). The other approaches evaluate locations that might be the optimum; for them we assume \(\bm{z}_t^\star = (\bm{x}^\star, \bm{\theta}^\star) = \argmax_{\bm{z} \in \mathcal{X} \times \Theta}\alpha(\bm{z} \vert D_t)\). 
Given these optima, we calculate regret measures. For problems with a discrete space of uncontrollable parameters \(\Theta\), where \(\bm{h}(\bm{x})\) is cheap to calculate, we directly take into account our robustness requirement by evaluating the robust regret \(\vert f(\bm{x}^\star, \bm{h}(\bm{x}^\star)) - f^\star \vert\). For problems with uncontrollable parameters from a continuous space \(\Theta\), such as the within-model comparison, \(\bm{h}(\bm{x})\) is hardly accessible. Therefore, we use the inference/immediate regret \(\vert f(\bm{z}_t^\star) - f^\star \vert\) for the evaluation of the \gls{RES} acquisition function/the other acquisition functions.
Notably, the metrics are non-monotonic, as the guess about the optimum can deteriorate with time. However, using a monotonic measure like best regrets, i.e., specifying the regret of the best found optimum up to iteration \(t\) for each run, is not helpful for min max problems. This is because pure minimization algorithms can find an optimum close to the robust optimum at the beginning of the optimization process and then converge to a non-robust optimum. The use of best regrets obscures this behavior. 

If not mentioned otherwise, we use a zero mean \gls{GP} prior, and a squared-exponential covariance function with automatic-relevance detection \(k(\bm{z}, \bm{z}') = \sigma_v^2 \exp \left(-0.5 \vert\vert \bm{z} - \bm{z}' \vert\vert^2_{\bm{L}^{-1}}\right)\) with \(\bm{L} = \text{diag}\left[l^2_{c1}, \dots, l^2_{d_c}, l^2_{u1}, \dots, l^2_{d_u} \right]\).

Runtime results for representative experiments are given in appendix~\ref{sec:practical_runtime}.

The code to conduct the experiments is built on open source implementations of \gls{GP}s \citep{gpy2014}, \gls{BO} \citep{emukit2023}, \gls{SSGP}s and \gls{EP} \citep{Frohlich2020} and publicly available at \url{https://github.com/fraunhofer-iais/Robust-Entropy-Search}.

\subsection{Within-Model Comparison}
\begin{figure*}
    \centering
    \includegraphics[width=\textwidth]{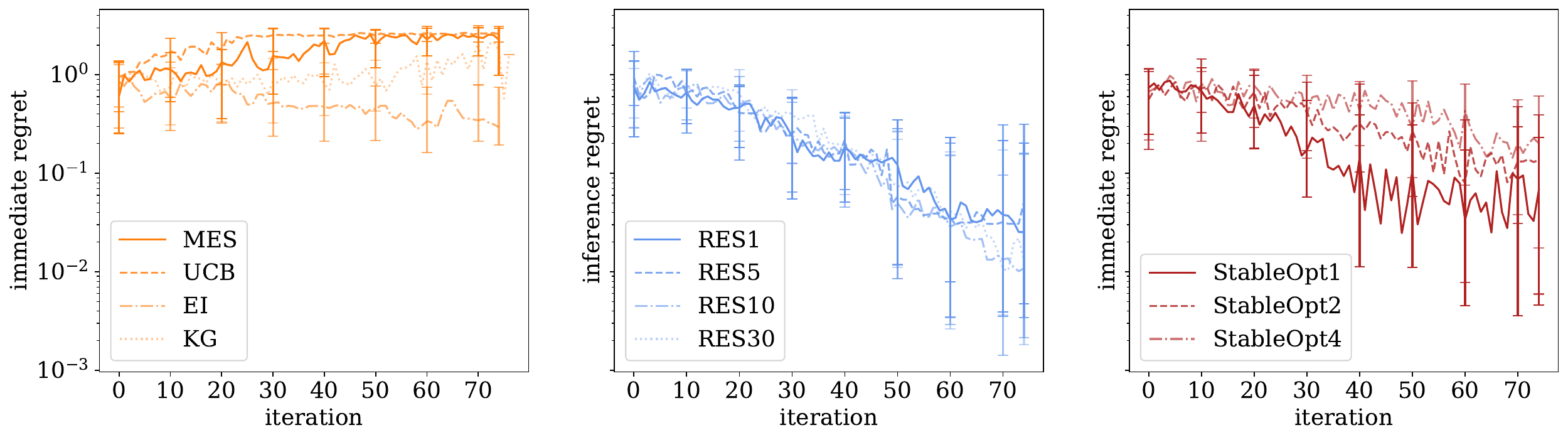}
    \caption{Regrets for the two-dimensional, continuous within-model comparison. We present the median and the upper and lower quartiles for 50 \gls{GP} mean functions. The number after the algorithm indicates the value of the hyperparameter (\(C\) for \gls{RES} and \(\sqrt{\beta}\) for \sopt). The results indicate the failure of the non-robust methods as well as the fact that \gls{RES} acquisition function is slightly better than \sopt{} with the advantage of being hyperparameter-free.}
    \label{fig:within_model}
\end{figure*}
For the within-model comparison, we follow the approach of \citet{Hennig2012} to compare the acquisition functions independently from the correct fit of the actual \gls{GP} model. 

Therefore, we use a \gls{GP} model with squared-exponential covariance function with signal variance \(\sigma_f^2 = 1\), a constant lengthscale of \(l = 0.1\) in all dimensions and a noise variance of \(\sigma_n^2 = 0.001\). 
For each of the 50 initializations, we
draw 1000 random data points whose locations follow a uniform distribution in \([0, 1]^2\) and whose values are distributed according to a normal distribution with zero mean and the covariance according to the specified covariance function. Given these points, we initialize a \gls{GP}.
Its predictive mean is employed as the objective function for optimization, so we deal with a two-dimensional continuous problem.
The motivating example in figure~\ref{fig:1} is one of the resulting optimization problems. For \gls{RES}, we apply numbers of function samples of \(C \in \lbrace1, 5, 10, 30 \rbrace\).

In figure~\ref{fig:within_model}, we report the results of the experiments.
As expected, the non-robust approaches are not able to find the robust optima.
For StableOpt, the performance on this particular problem depends on the value of the exploration parameter - (for the within-model comparison) the lower, the better. Generally, our approach \gls{RES} is better than \sopt{} and, advantageously, does not require setting a hyperparameter. 
Additionally, the number of samples only slightly impacts the performance of \gls{RES}. Therefore, we set the number of samples to 1 for all other experiments.

\subsection{Synthetic Benchmark Functions}
\begin{figure*}[h]
\centering
    \begin{subfigure}[t]{0.3\textwidth}
    \includegraphics[width=\textwidth]{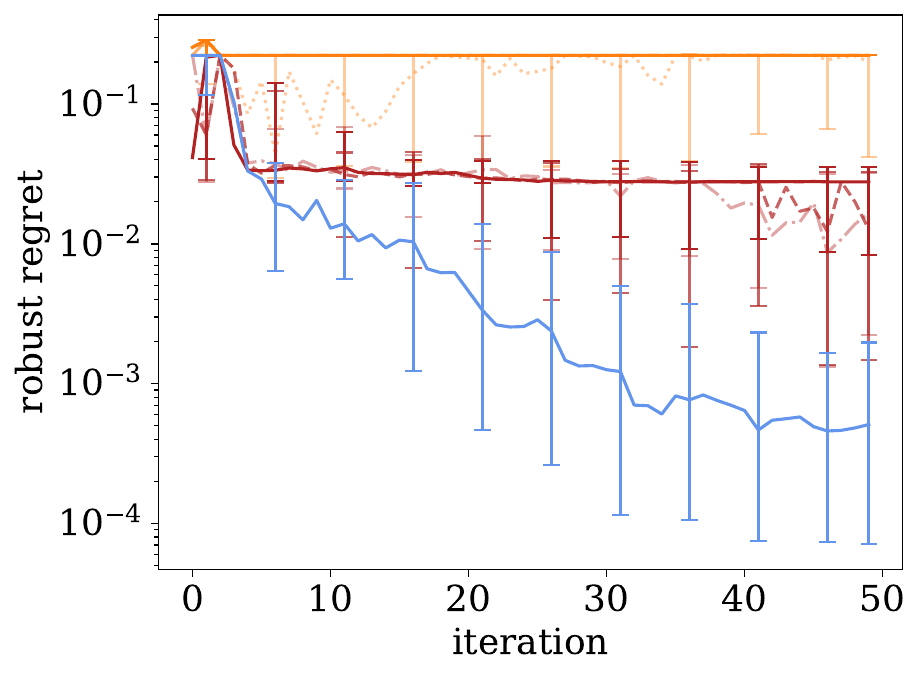}
    \caption{Branin (2 d., \(\bm{x}\): conti., \(\vert\bm{\theta}\vert = 20\))}
    \label{fig:first}
\end{subfigure}
\hfill
\begin{subfigure}[t]{0.3\textwidth}
    \includegraphics[width=\textwidth]{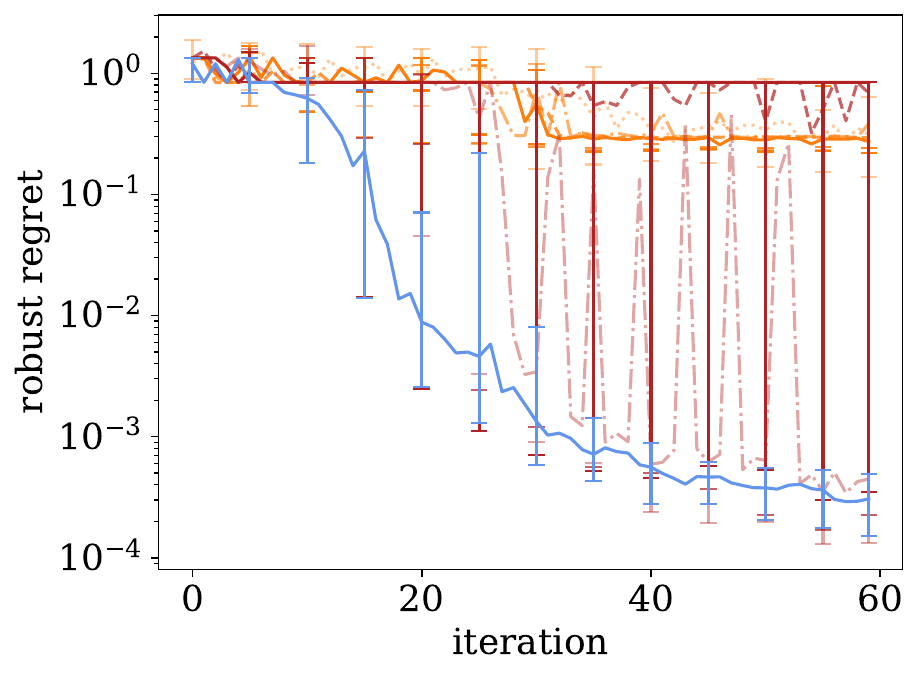}
    \caption{Sinus + Linear (2 d., \(\bm{x}\): conti., \(\vert\bm{\theta}\vert = 2\))}
    \label{fig:second}
\end{subfigure}
\hfill
\begin{subfigure}[t]{0.3\textwidth}
    \includegraphics[width=\textwidth]{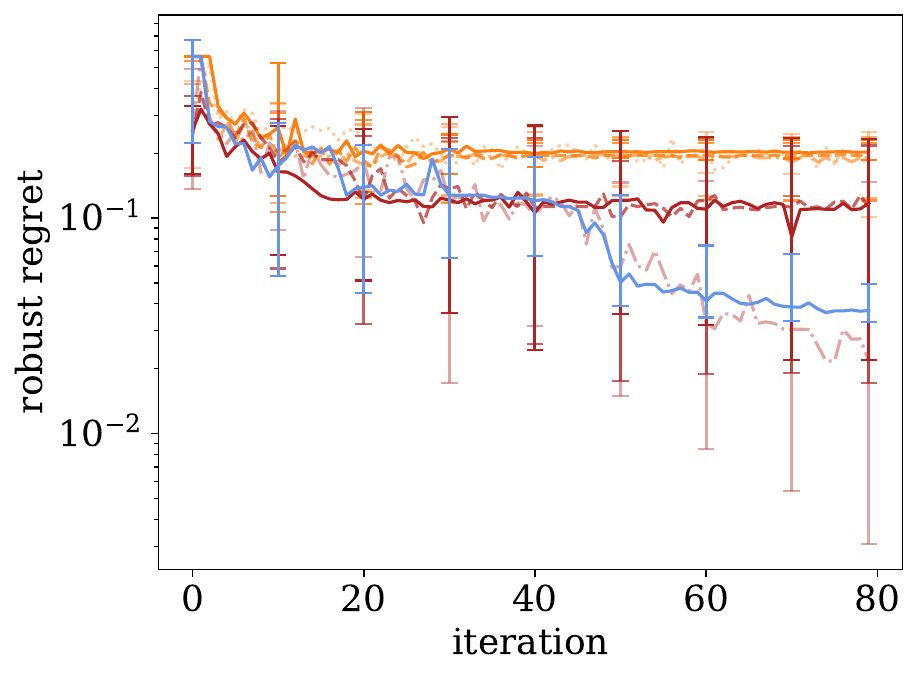}
    \caption{ Eggholder (2 d., \(\bm{x}\): conti., \(\vert\bm{\theta}\vert = 3\))}
    \label{fig:third}
\end{subfigure}
\hfill
\begin{subfigure}[t]{0.3\textwidth}   
    \centering 
    \includegraphics[width=\textwidth]{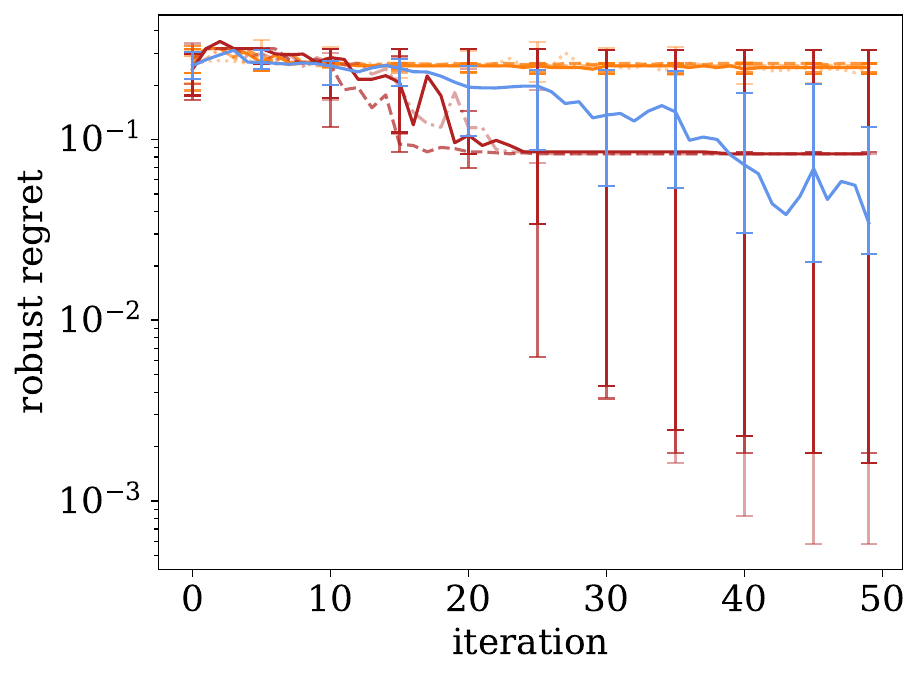}
    \caption{Hartmann (3 d., \(\vert\bm{x}\vert = 2500, \vert\bm{\theta}\vert = 11\))}%
    \label{fig:gmm}
\end{subfigure}
\hfill
\begin{subfigure}[t]{0.3\textwidth}   
    \centering 
    \includegraphics[width=\textwidth]{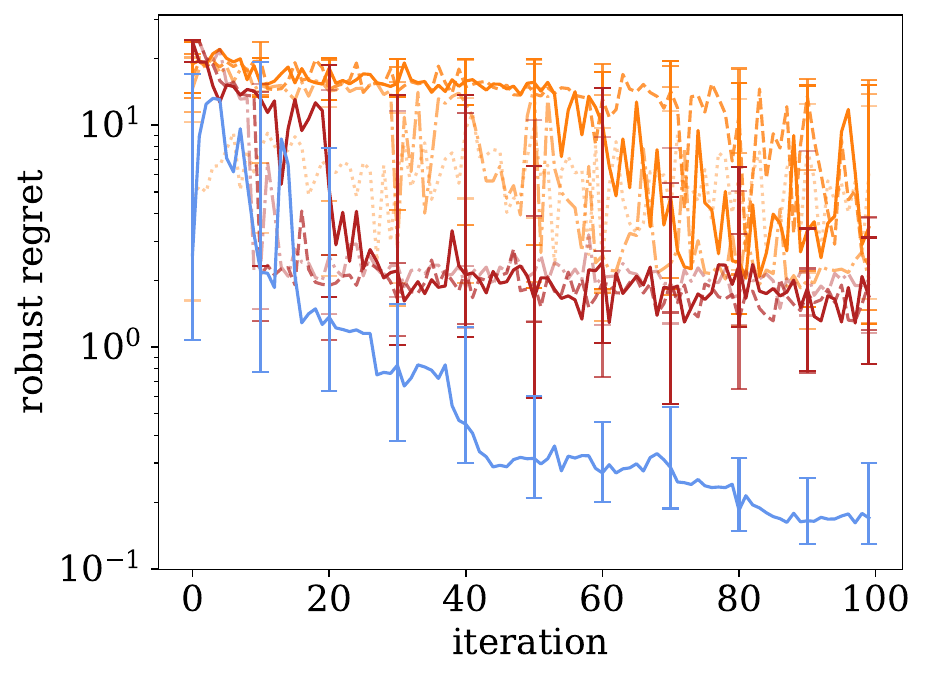}
    \caption{Synthetic Polynomial (4 d., \(\bm{x}\): 2 d., conti., \(\vert \bm{\theta} \vert = 12\))}%
    \label{fig:synth_poly}
\end{subfigure}
\hfill
\begin{subfigure}[t]{0.3\textwidth}   
    \centering 
    \includegraphics[width=0.5\textwidth]{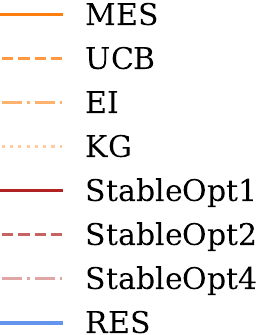}
    \label{fig:legend}
\end{subfigure}
\caption{Results of the experiments with synthetic functions. The marker after the name of the problem indicates the dimensionality, the type of input space (continuous or discrete) and, if discrete, the number of discrete parameters. For \sopt{}, we give the value of the exploration constant \(\sqrt{\beta}\) after the algorithm name. Our approach, \gls{RES}, with a number samples of \(C =1\), shows superior performance on nearly all problems.}
\label{fig:synthetic_problems}
\end{figure*}
In the synthetic experiments, we measure the performance of our approaches on problems with unknown hyperparameters.
Basically, we use variations of the Branin \citep{optimization_functions}, the Sinus + Linear \citep{Frohlich2020}, the Eggholder \citep{optimization_functions}, the Hartmann 3D \citep{optimization_functions}, and the Synthetic Polynomial \citep{Bertsimas2010NonconvexRO}
functions. In these originally non-robust optimization problems, we declare a subset of dimensions as uncontrollable parameters \(\bm{\theta}\) and then search for the robust optimum.

To reduce the computational effort, we discretize the space of the uncontrollable parameters \(\Theta\) in all experiments. The exact number of uncontrollable parameters is given below the figures, as well as the problem's dimensionality. 
Full details and visualizations of the individual problems are given in appendix~\ref{sec:Synthetic_Experiments}.

We run each algorithm with 50 initializations. 
For all problems, except from the Synthetic Polynomial where we fix the hyperparameters, we optimize the hyperparameters of the \gls{GP} model in every iteration via maximum likelihood. The noise hyperparameter \(\sigma_n\) is fixed to a value of \(0.001\) in all problems. 

In figure~\ref{fig:synthetic_problems}, we report the performance of all algorithms in terms of the quartiles.
The experiments show a superior performance of \gls{RES} over the other approaches, independent from the dimensionality or the complexity of the problem (e.g., the eggholder problem having a lot of local optima). In some cases, algorithms oscillate between different optima, i.e., for the \sopt{} algorithm with \(\sqrt{\beta} = 4\) in the Sinus + Linear problem and for the non-robust algorithms in the Synthetic Polynomial. This is due to the very different values of the max function \(g(\bm{x})\) for different inputs \(\bm{x}\).
Additionally, the previously en par \sopt{} algorithm struggles with the fixed or unknown hyperparameters of the model. This behavior was already reported for plain \gls{UCB} in \cite{Hennig2012} and seems to apply also for the robust adaption. Also, due to the unknown hyperparameters, \sopt{} underlies the risk of too early exploitation. In these cases, one of the local robust optima is preferred over the global one, increasing the width of the distribution over results. Therefore, \sopt{} often reaches better results in the lower quantiles (if it examines the correct local optimum). However, its median behavior is worse than that of our approach, as RES is forced to explore more globally as it has to learn not only about the robust optimum but also about the other robustness characteristics. This behavior becomes particularly clear in the Sinus + Linear, the Eggholder and the Hartmann problems.

In appendix~\ref{sec:performance_runtime} we additionally provide results on the robust regret over the runtime for the Branin function. \gls{RES} achieves a similar regret in the same time as \sopt{} with a significantly lower number of iterations.

\subsection{Real-Life Benchmark Problems}
We treat two benchmark problems connected to applying robust \gls{BO} in real life. 

\subsubsection{Calibration of Finite Element Method Simulation Parameters}
\begin{figure*}
    \centering
    \begin{subfigure}[t]{0.3\textwidth}
        \centering
        \includegraphics[width=\textwidth]{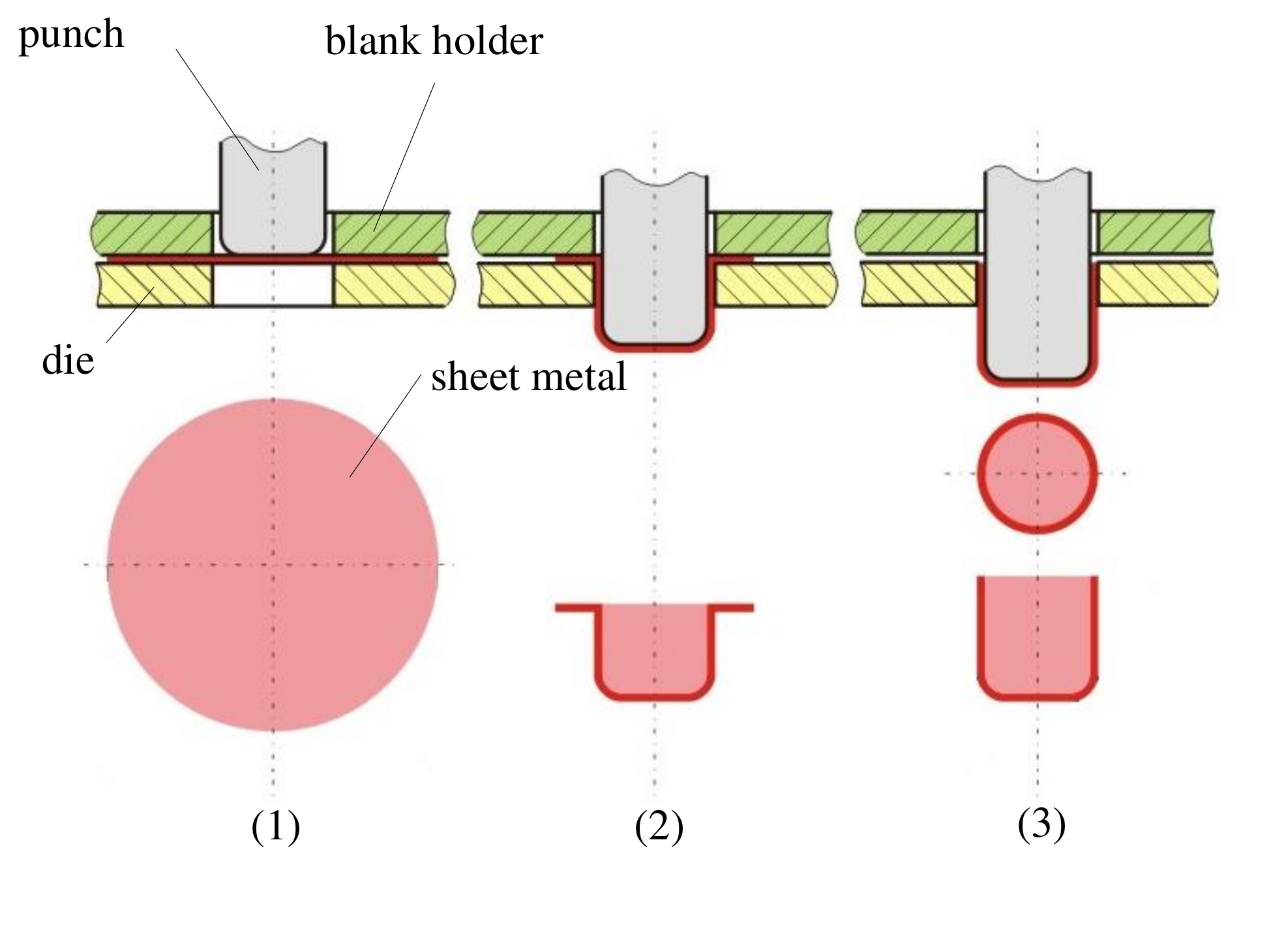}
        \caption{Deep Drawing Process \citep{figdeepdrawing}}%
        \label{fig:deep_drawing}
    \end{subfigure}
    \hfill
    \begin{subfigure}[t]{0.3\textwidth}  
        \centering 
        \includegraphics[width=\textwidth]{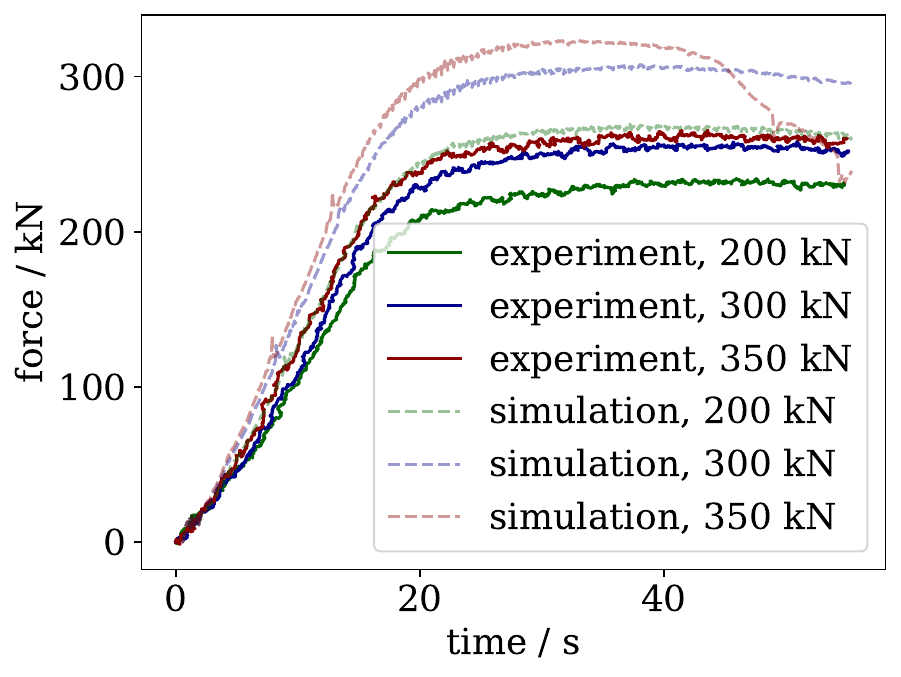}
        \caption{Simulated and experimental results of Force-Time Diagram for \(\mu_H = 0.2\).}%
        \label{fig:force-timediagram}
    \end{subfigure}
    \hfill
    \begin{subfigure}[t]{0.3\textwidth}  
        \centering 
        \includegraphics[width=\textwidth]{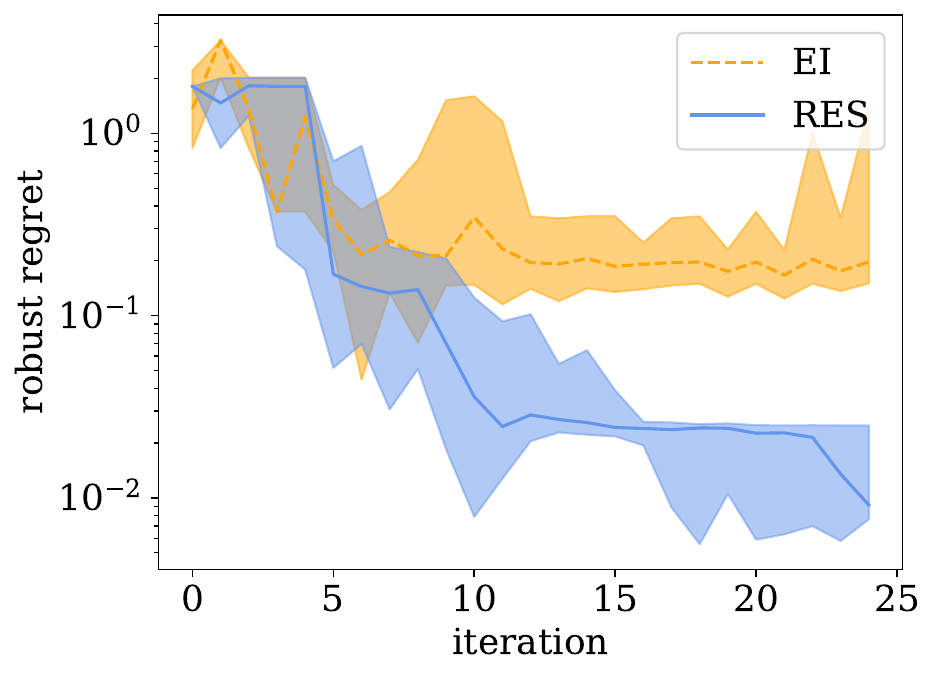}
        \caption{Robust Regret for \gls{RES} and \gls{EI} acquisition functions.}%
        \label{fig:FEM_calibration}
    \end{subfigure}
    \caption{Deep drawing: schematic illustration, force-time-diagrams and regret curves for simulations with different parameters.}%
    \label{fig:real_life_experiments}
\end{figure*}
In engineering disciplines, a lot of research and development tasks involve the application of heavy simulations, e.g., simulations via Finite Element Method.
These simulations are typically taking minutes to months for execution. Nevertheless, they are advantageous over real-life experiments in the lab, as they are often cheaper (they do not induce, e.g., material costs) and enable insights of multiple metrics at each time step and many locations simultaneously.
Unfortunately, the approximation quality of simulations depends on material parameters, which are typically unknown. These parameters are not directly measurable and depend on exogenous parameters, such as local temperature. Therefore, a set of experiments is taken out at a range of different uncontrollable parameters, and engineers use the result to calibrate the simulations, i.e., to fit the unknown (controllable) material parameters to approximate the experiments.

In our use case, we treat the calibration of simulation parameters of a deep drawing process, where one simulation takes about 12 minutes on 16 cores of an Intel(R) Core(TM) i9-10980XE processor. In deep drawing, a metal sheet is placed on a die, held in place by a blank holder, and drawn into a new shape by pressing a punch, see figure~\ref{fig:deep_drawing}. Experimentally, the force of the punch \(F_{\text{punch}_{\text{ex}}}\) was measured over time, varying the constant force of the blank-holder \(F_{\text{holder}} \in \lbrace 200, 300, 350 \rbrace\)~kN. The static coefficient of friction \(\mu_H \in \left[0.1, 0.2 \right]\) is treated as the controllable parameter, which depends, as no lubrication is used, only on the (unknown) surface quality of the die, punch, blank holder, and the metal sheet. 
Exemplary experimental and simulated force-time diagrams are shown in figure~\ref{fig:force-timediagram}.
The optimization objective is to minimize the maximum absolute difference between the experimental and simulated punch force, so we seek to find \(\mu_H^\star = \argmin_{\mu_H} \max_{F_{\text{Holder}}} \vert F_{\text{punch}_{\text{ex}}} - F_{\text{punch}_{\text{sim}}} \vert = \argmin_{\mu_H} \max_{F_{\text{holder}}} f(\mu_H, F_{\text{Holder}})\). 

We run our optimization approach 30 times for 25 iterations for the \gls{RES} and the \gls{EI} acquisition function, each with one random sample for initialization. Hyperparameters of the model are estimated in every iteration via maximum likelihood. To find the robust optimum for comparison, we join the data from all 750 evaluations, create a \gls{GP} model, and calculate the function value at the model's optimum.
Figure~\ref{fig:FEM_calibration} shows the optimization results in terms of robust regret: while \gls{EI} soon finds some non-robust optimum, \gls{RES} finds a considerably better robust optimum already after ten iterations. While more iterations would have been interesting from a scientific perspective, the results were already sufficient for the application side. The robust optimal coefficient of friction \(\mu_H^\star\) is now used as a safe estimate for simulations with unknown blank-holder force.

\subsubsection{Robust Robot Pushing}
In appendix~\ref{sec:robot_pushing}, we provide results on the robust robot pushing problem from \cite{Bogunovic} and find \gls{RES} again performing best.

\section{Conclusion}
We introduced a novel worst-case robust acquisition function for \gls{BO}, \gls{RES}. In a nutshell, this acquisition function simultaneously maximizes the information gain about the robust objective function \(g\), the location of the robust objective function \(\bm{h}\), and the robust optimal value \(f^\star\). In several benchmark experiments, we demonstrate the superior efficiency of our acquisition function and show its benefit in two use cases from engineering and robotics.

\section{Limitations and Future Work}
This paper's main contribution is developing an innovative information-theoretic acquisition function for adversarially robust \gls{BO}. When used with a sufficiently accurate model, it produces impressive results.
However, its performance relies on the correctness of the model, which is not necessarily the case for complex problems. 
A simple technique to detect a poor model fit is via the \(\gamma\)-exploit approach, used by \citet{Hvarfner2022}, where in each iteration, with probability \(\gamma\), the actual optimum is evaluated. Unfortunately, this approach detects but does not circumvent a poor model fit. 
Therefore, we expect an even more significant improvement in combination with automatic model selection methods, such as those by \citet{Malkomes2018,Gardner2019}. Especially the ability to discover additive structures in the work of \citet{Gardner2019} promises to additionally scale the approach to higher-dimensional spaces, thus being a valuable enhancement.

Additionally, the derivation of regret bounds would likewise be interesting, such as typically performed in \gls{UCB}-based approaches, such as the \sopt{} algorithm \citep{Bogunovic}. For information-based acquisition functions, we are only aware of the disputed \citep{Takeno2022} regret bounds for \gls{MES} and its descendants \citep{Wang17,Belakaria2019}. An extension of the existing work, considering the recent discussions and the robust setting of our approach, is a challenging open problem.

For future work, we intend to adapt our algorithm to various domains, such as the constrained \citep{gelbart2014,Gardner2014}, the multi-fidelity \citep{Forrester2007MultifidelityOV}, and multi-objective \citep{Swersky2013} setting.

\begin{contributions} 
    D. Weichert conceived the idea of the paper together with A. Kister, created the code, the figures, wrote the initial draft of the manuscript and performed revisions.
    A. Kister conceived the idea of the paper and revised the manuscript.
    P. Link performed the simulations via Finite Element Method.
    S. Houben revised the manuscript.
    G. Ernis revised the initial draft of the manuscript.
\end{contributions}

\begin{acknowledgements} 
We thank the reviewers for their helpful feedback.
The work of D. Weichert has been funded by the Federal Ministry of Education and Research of Germany and the state of North Rhine-Westphalia as part of the Lamarr Institute for Machine Learning and Artificial Intelligence, Sankt Augustin, Germany.
P. Link was funded by the Deutsche Forschungsgesellschaft (DFG, German Research Foundation) - 438646126.
\end{acknowledgements}

\bibliography{literature}

\newpage

\onecolumn

\title{Adversarially Robust Entropy Search for Safe Efficient Bayesian Optimization}
\maketitle
\appendix
\section{Approximation of Bivariate Doubly Truncated Gaussian}
\label{sec:doubly_truncated}
We closely follow the results of \cite{ang2002}.  Let \(\bm{x} = (x_1, x_2) \sim \mathcal{N}(\bm{0}, \bm{\Sigma})\), with lower bounds \(\bm{l}_b = \begin{bmatrix} l_{b_1} & l_{b_2} \end{bmatrix}^T\) and upper bounds \(\bm{u}_b = \begin{bmatrix} u_{b_1} & u_{b_2}\end{bmatrix}^T\), and \(\rho\) denote the correlation of the two variables.

Let the cumulative density be denoted by \(L\)
\begin{equation*}
    L(\bm{l}_b, \bm{u}_b) = \int_{l_{b_1}}^{u_{b_1}} \int_{l_{b_2}}^{u_{b_2}} f_{\bm{x}}(x_1, x_2) dx_1 dx_2 ~,
\end{equation*}
with \(f_{\bm{x}}(x_1, x_2)\) being the density function of \(\bm{x}\).
\(L\) can be evaluated numerically, e.g., using the method of \cite{Genz1992}.

For the moments, we find:
\begin{equation}
    m_{10} = \frac{1}{L} \left[\psi(l_{b_1}, u_{b_1}, l_{b_2}, u_{b_2}) + \rho \psi(l_{b_1}, u_{b_1}, l_{b_2}, u_{b_2}) \right] 
\end{equation}
\begin{equation}
    m_{20} =  \frac{1}{L} \left[L + \chi(l_{b_2}, u_{b_2}, l_{b_1}) - \chi(l_{b_2}, u_{b_2}, u_{b_1}) + \rho^2 \chi(l_{b_1}, u_{b_1}, l_{b_2}) - \rho^2 \chi(l_{b_1}, u_{b_1}, u_{b_2}) \right]
\end{equation}
\begin{equation}
\begin{split}
    m_{11} &= \frac{1}{L} \left[ \rho L + \rho \Upsilon(l_{b_1}, u_{b_1}, l_{b_2}) - \rho \Upsilon(l_{b_1}, u_{b_1}, u_{b_2}) + \rho \Upsilon(l_{b_2}, u_{b_2}, l_{b_1}) - \rho \Upsilon(l_{b_2}, u_{b_2}, u_{b_1}) \right. \\
    &+ \left. \Lambda(l_{b_1}, u_{b_1}, l_{b_2}) - \Lambda(l_{b_1}, u_{b_1}, u_{b_2}) \right]
\end{split}
\end{equation}

with helper functions 
\begin{equation*}
    \psi(l_{b_1}, u_{b_1}, l_{b_2}, u_{b_2}) = \phi(l_{b_1}) \left[\Phi \left(\frac{u_{b_2} - \rho l_{b_1}}{\sqrt{1-\rho^2}} \right) - \Phi\left(\frac{l_{b_2} - \rho l_{b_1}}{\sqrt{1-\rho^2}}\right) \right] - \phi(u_{b_1}) \left[\Phi \left(\frac{u_{b_2} - \rho u_{b_1}}{\sqrt{1-\rho^2}} \right) - \Phi \left(\frac{l_{b_2} - \rho u_{b_1}}{\sqrt{1-\rho^2}} \right) \right]~,
\end{equation*}

\begin{equation*}
    \begin{split}    
        \chi(l_{b_2}, u_{b_2}, l_{b_1}) &= l_{b_1} \phi(l_{b_1}) 
        \left[\Phi \left( \frac{u_{b_2} - \rho l_{b_1}}{\sqrt{1-\rho^2}} \right) - \Phi \left( \frac{l_{b_2} - \rho l_{b_1}}{\sqrt{1-\rho^2}} \right) \right] \\ 
        &+ \frac{\rho \sqrt{1-\rho^2}}{\sqrt{2 \pi} \left(1 + \rho^2 \right)}
        \left[\phi \left( \frac{\sqrt{l_{b_2}^2 - 2 \rho l_{b_2} l_{b_1} + l_{b_1}^2}}{\sqrt{1 - \rho^2}} \right) - \phi \left( \frac{\sqrt{u_{b_2}^2 - 2 \rho u_{b_2} l_{b_1} + l_{b_1}^2}}{\sqrt{1 - \rho^2}} \right) \right]~,
    \end{split}
\end{equation*}

\begin{equation*}
    \Upsilon(l_{b_2}, u_{b_2}, l_{b_1}) = l_{b_1} \phi(l_{b_1}) \left[ \Phi \left( \frac{u_{b_2}- \rho l_{b_1}}{\sqrt{1-\rho^2}} \right) - \Phi \left( \frac{l_{b_2} - \rho l_{b_1}}{\sqrt{1-\rho^2}} \right) \right]~, 
\end{equation*}
and
\begin{equation*}
    \Lambda(l_{b_2}, u_{b_2}, l_{b_1}) = \frac{\sqrt{1 - \rho^2}}{\sqrt{2 \pi}} \left[ \phi \left( \frac{\sqrt{l_{b_2}^2 - 2 \rho l_{b_2} l_{b_1} + l_{b_1}^2}}{\sqrt{1-\rho^2}} \right) - \phi \left( \frac{\sqrt{u_{b_2}^2 - 2 \rho u_{b_2} l_{b_1} + l_{b_1}^2}}{\sqrt{1-\rho^2}} \right) \right]~,
\end{equation*}

where \(\phi\) is the probability density function and \(\Phi\) is the cumulative density function of the standard normal \(\mathcal{N}(0, 1)\). 

The moments \(m_{01}\) and \(m_{02}\) are obtained by interchanging \((l_{b_1}, u_{b_1})\) and \((l_{b_2}, u_{b_2})\) in the formulae for \(m_{10}\) and \(m_{20}\).

Given these moments, we finally find the following approximating normal distribution \(\mathcal{N}(\hat{\bm{\mu}}, \hat{\bm{\Sigma}})\) with 
\(
    \hat{\bm{\mu}} = \begin{bmatrix}
        m_{10} & m_{01}
    \end{bmatrix}^T
\) and
\(
    \hat{\bm{\Sigma}} = \begin{bmatrix}
        m_{20} - m_{10}^2 & m_{11} - m_{10} m_{01} \\
        m_{11} - m_{10} m_{01} & m_{02} - m_{01}^2
    \end{bmatrix}
\).
From these, we extract \(m_q = m_{10}\) and \(v_q = m_{20} - m_{10}^2\).

\section{Detailed Description of Experiments with Synthetic Benchmark Functions}
\label{sec:Synthetic_Experiments}

\paragraph{Branin Function}
The branin function is defined by
\begin{equation*}
    f(\bm{x}, \bm{\theta}) = a(\bm{\theta} - b \bm{x}^2 + c\bm{x} -r)^2 + s (1-t) \cos(\bm{x}) + s ~,
\end{equation*}
with \(a = 1\), \(b = 5.1/(4 \pi^2)\), \(c = 5 / \pi\), \(r = 6\), \(s = 10\), and \(t = 1 / (8 \pi)\) and is defined on \(x \in \left[-5, 10 \right]\), \(\theta \in \left[0, 15 \right]\) \citep{optimization_functions}.

We use discrete values of the uncontrollable parameter \(\theta \in \lbrace0.75, 1, \dots, 14, 14.25 \rbrace, \vert \Theta \vert = 20 \), and scale the input space to \(\left[0, 1 \right]^2\) and the output values to \(\mathcal{N}(0, 1)\).

For optimization, the hyperparameters of the \gls{GP} are bounded to \(\sigma_v \in \left[10^{-5}, 10\right]\) and \(\bm{l} \in \left[10^{-5}, 10\right]^2\). The model is initialized with a single random point from the domain. We run each algorithm with 50 different initializations for 50 iterations.

\begin{figure}[h]
    \centering
    \begin{subfigure}[b]{0.45\textwidth}
        \centering
        \includegraphics[width=\textwidth]{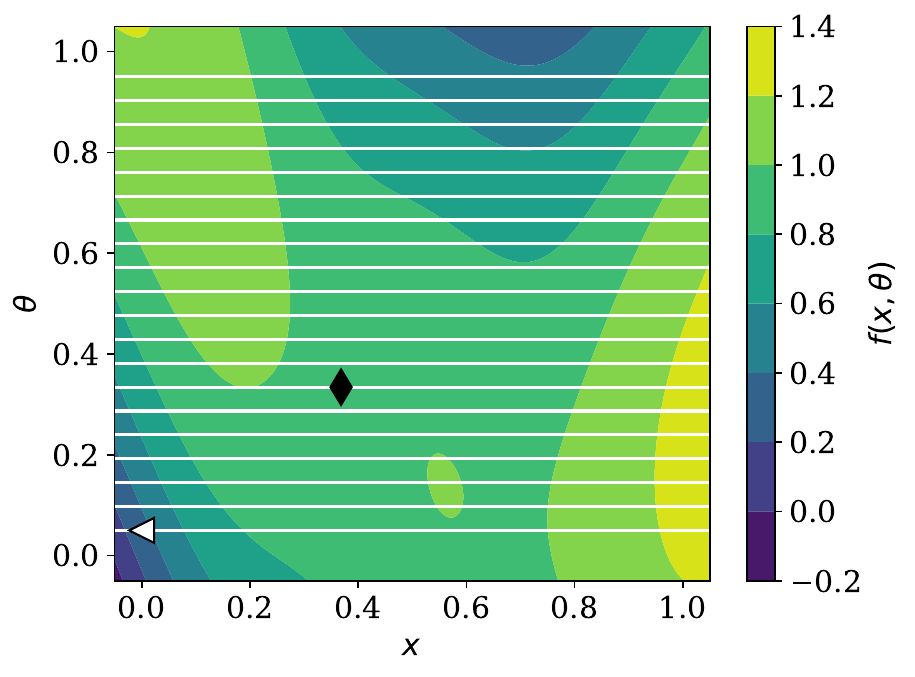}
        \caption{Branin Function}%
        \label{fig:branin1}
    \end{subfigure}
    \hfill
    \begin{subfigure}[b]{0.45\textwidth}
        \centering
        \includegraphics[width=\textwidth]{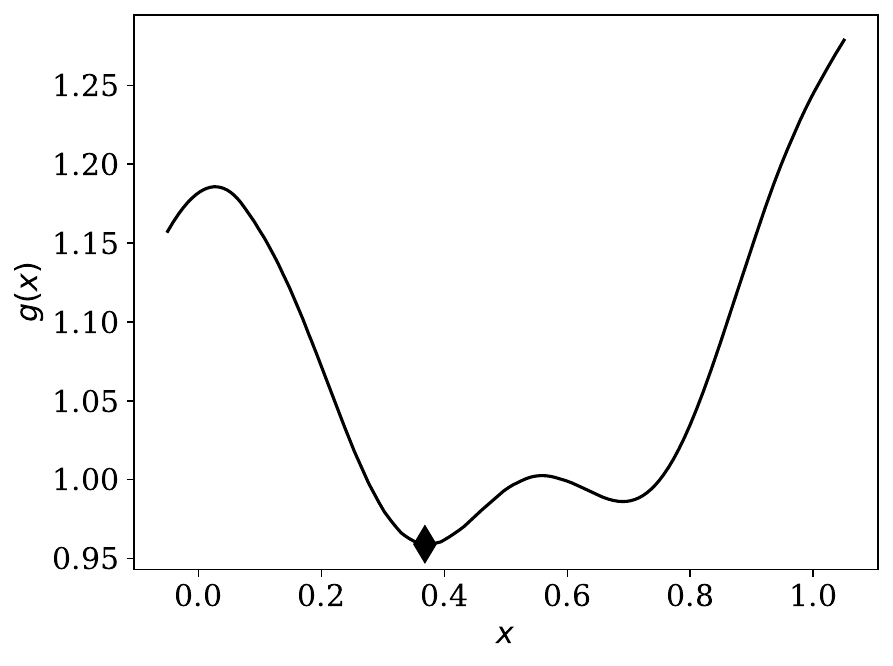}
        \caption{Robust Branin Function}%
        \label{fig:branin2}
    \end{subfigure}
    \caption{Visualization of the robust variant of Branin Function. The global robust optimum is indicated by $\blacklozenge$, the global minimum by $\lhd$.}%
    \label{fig:branin}
\end{figure}

Figure~\ref{fig:branin1} shows the original optimization problem with the 20 discrete values of the uncontrollable parameter as white horizontal lines, the robust optimum and the global minimum. The maximizing function \(g(\bm{x}) = \max_{\bm{\theta}} f(\bm{x}, \bm{\theta})\) is visualized in figure~\ref{fig:branin2}.

\paragraph{Sinus + Linear Function}
The sinus + linear function is defined by 
\begin{equation*}
    f(\bm{z}) = \sin{(5 \bm{z}^2 \pi)} + 0.5 \bm{z}~,
\end{equation*}
where \(\bm{z} = \bm{x} + \bm{\theta}\) with \(x \in \left[0, 1\right]\) and \(\theta \in \lbrace 0.1, 0.05 \rbrace\). It was originally used by \cite{Frohlich2020} with continuous \(\theta \in \left[-0.05, 0.05\right]\). We opted for discretization for the sake of simplicity.

\begin{figure}[h]
    \centering
    \begin{subfigure}[b]{0.45\textwidth}
        \centering
            \includegraphics[width=\textwidth]{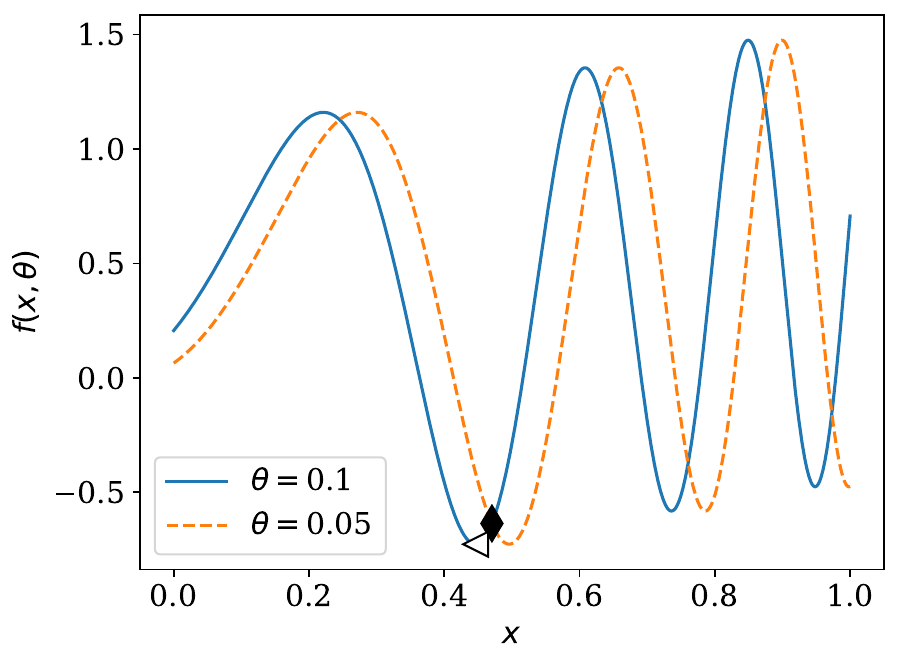}  
        \caption{Sinus + Linear Function}%
        \label{fig:sin_p_linear}
    \end{subfigure}
    \hfill
    \begin{subfigure}[b]{0.45\textwidth}
        \centering
        \includegraphics[width=\textwidth]{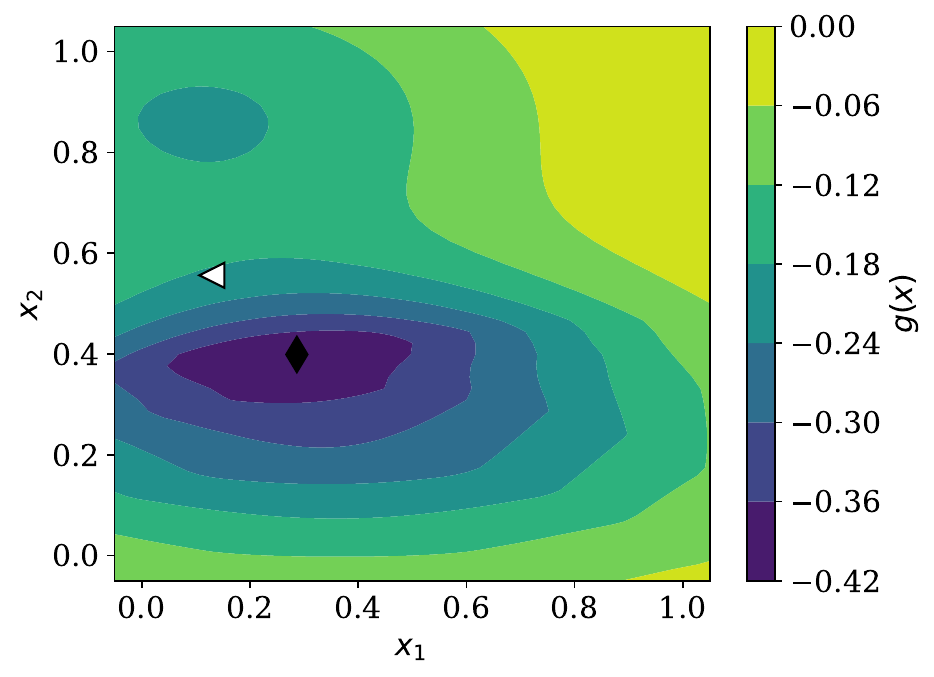}  
        \caption{Robust Hartmann 3d function}%
        \label{fig:hartmann2}
    \end{subfigure}
    \caption{Visualization of robust Sinus + Linear and Hartmann function variants. The global robust optimum is indicated by $\blacklozenge$ and the global minimum by $\lhd$.}
\end{figure}

Figure~\ref{fig:sin_p_linear} visualizes the problem. Multiple local robust and non-robust optima exist, which are close to each other.

For optimization, the hyperparameters of the \gls{GP} are bounded to \(\sigma_v \in \left[10^{-5}, 10\right]\) and \(\bm{l} \in \left[10^{-5}, 10\right]^2\). The model is initialized with a single random point from the domain.
We run each algorithm with 50 different initializations for 60 iterations.

\paragraph{Hartmann Function}
Following \cite{optimization_functions}, the three-dimensional Hartmann function is defined by 
\begin{equation*}
    f(\bm{z}) = \sum_{i=1}^4 \alpha_i \exp{\left(-\sum_{j=1}^3 A_{ij} \left(z_j - P_{ij}\right)^2 \right)}~,
\end{equation*}
where \(\alpha = \begin{bmatrix}
    1.0 & 1.2 & 3.0 & 3.2
\end{bmatrix}^T\), 
\(\bm{A} = \begin{bmatrix}
    3 & 10 & 30 \\ 0.1 & 10 & 35 \\ 3 & 10 & 30 \\ 0.1 & 10 & 35 
\end{bmatrix}\),
\(\bm{P} = 10^{-4} \begin{bmatrix}
    3689 & 1170 & 2673 \\ 4699 & 4387 & 7470 \\ 1091 & 8732 & 5547 \\ 381 & 5743 & 8828
\end{bmatrix}\).
It is defined on \(\bm{z}  \in \left[0,1\right]^3\). In our experiments, we use the first two dimensions as controllable parameters \(\bm{x}\) on an equidistant grid of size \(50 \times 50 = 2500\), and use the third dimension as uncontrollable parameter \(\theta\), which is discretized to values of \(\lbrace0.25, 0.3,\dots, 0.7, 0.75 \rbrace, \vert\Theta\vert = 11\).

The maximizing function \(g(\bm{x}) = \max_{\bm{\theta}} f(\bm{x}, \bm{\theta})\), the robust optimum and the global minimum are visualized in figure~\ref{fig:hartmann2}.

For optimization, the hyperparameters of the \gls{GP} are bounded to \(\sigma_v \in \left[10^{-5}, 10\right]\) and \(\bm{l} \in \left[10^{-5}, 10\right]^2\). The model is initialized with a single random point from the domain.
We run each algorithm with 100 different initializations for 50 iterations.

\paragraph{Eggholder Function}
Following \cite{optimization_functions}, the eggholder function is defined by
\begin{equation*}
    f(\bm{x}, \bm{\theta}) = -(\bm{\theta} + 47) \sin{\left(\sqrt{\left\vert \bm{\theta} + \frac{\bm{x}}{2} + 47\right\vert}\right) - \bm{x} \sin{\left(\sqrt{\lvert \bm{x} - (\bm{\theta} + 47) \rvert}\right)}}~,
\end{equation*}
with \(x \in \left[-512, 512 \right]\), \(\theta \in \left[-512, 512 \right]\).

We use discrete values of the uncontrollable parameter \(\theta \in \lbrace-512, 0, 185 \rbrace \), and scale the input space to \(\left[0, 1 \right]^2\) and the output values to zero mean and a variance of 1.

\begin{figure}[h]
    \centering
    \begin{subfigure}[t]{0.45\textwidth}
        \centering
        \includegraphics[width=\textwidth]{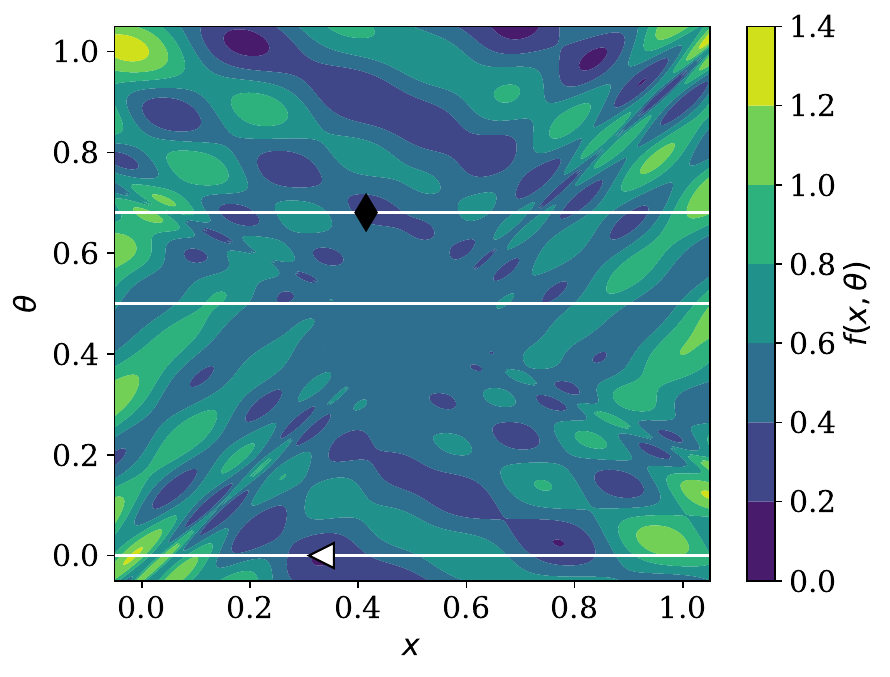}
        \caption{Eggholder Function}%
        \label{fig:eggholder1}
    \end{subfigure}
    \hfill
    \begin{subfigure}[t]{0.45\textwidth}
        \centering
        \includegraphics[width=\textwidth]{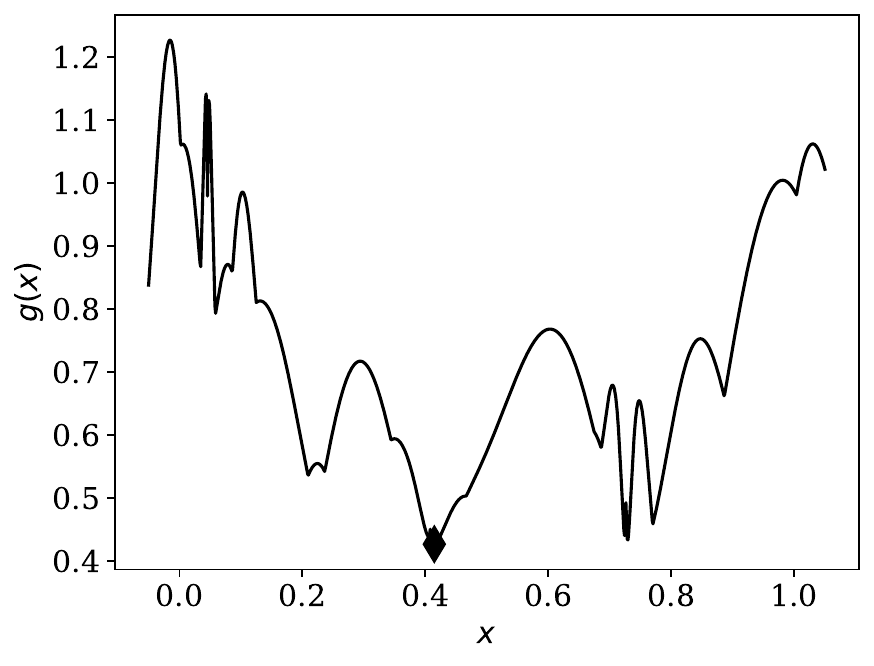}
        \caption{Robust Eggholder Function}%
        \label{fig:eggholder2}
    \end{subfigure}
    \caption{Visualization of the robust variant of Eggholder Function. The global robust optimum is indicated by $\blacklozenge$, the global minimum by $\lhd$.}%
    \label{fig:eggholder}
\end{figure}

Figure~\ref{fig:eggholder1} shows the original optimization problem with the three uncontrollable parameters as white horizontal lines, as well as the robust optimum. The maximizing function \(g(\bm{x}) = \max_{\bm{\theta}} f(\bm{x}, \bm{\theta})\) is visualized in figure~\ref{fig:eggholder2}.

For optimization, the hyperparameters of the \gls{GP} are bounded to \(\sigma_v \in \left[10^{-5}, 10\right]\) and \(\bm{l} \in \left[10^{-5}, 10\right]^2\).
The model is initialized with a single random point from the domain.
We run each algorithm with 50 different initializations for 80 iterations.

\paragraph{Synthetic Polynomial}
We adopt the synthetic polynomial, which has already been considered in multiple variations by \citet{Bertsimas2010NonconvexRO,Bogunovic,Frohlich2020,Christianson2023}.
It is originally defined by \cite{Bertsimas2010NonconvexRO}:
\begin{equation*}
    \begin{split}
        f(\bm{z}) = 2 \bm{z}_1^6 - 12.2\bm{z}_1^5 + 21.2 \bm{z}_1^4 + 6.2\bm{z}_1 -6.4 \bm{z}_1^3 - 4.7\bm{z}_1^2 \\
        + \bm{z}_2^6 - 11\bm{z}_2^5 + 43.3 \bm{z}_2^4 - 10\bm{z}_2 -74.8\bm{z}_2^3 + 56.9\bm{z}_2^2 \\
        -4.1 \bm{z}_1 \bm{z}_2 - 0.1 \bm{z}_2^2 \bm{z}_1^2 + 0.4 \bm{z}_2^2 \bm{z}_1 + 0.4 \bm{z}_1^2 \bm{z}_2
    \end{split}
\end{equation*}
with \(\bm{z} = \begin{bmatrix}
    \bm{z}_1 & \bm{z}_2
\end{bmatrix}\). We choose \(\bm{x}_1 \in \left[-0.95, 3.2\right]\) and \(\bm{x}_2 \in \left[-0.45, 4.4\right]\), and \(\bm{\theta}\) in a circular neighborhood with radii \(r \in \lbrace 0, 0.5 \rbrace\) and angles \(\alpha \in \lbrace0, 0.4\pi, 0.8\pi, 1.2\pi, 1.6\pi, 2\pi \rbrace\), so \(\bm{z} = \bm{x} + \bm{\theta} = \bm{x} + r \begin{bmatrix}
    \cos \alpha & \sin \alpha
\end{bmatrix}\). 

\begin{figure}[h]
    \centering
    \begin{subfigure}[t]{0.45\textwidth}
        \centering
        \includegraphics[width=\textwidth]{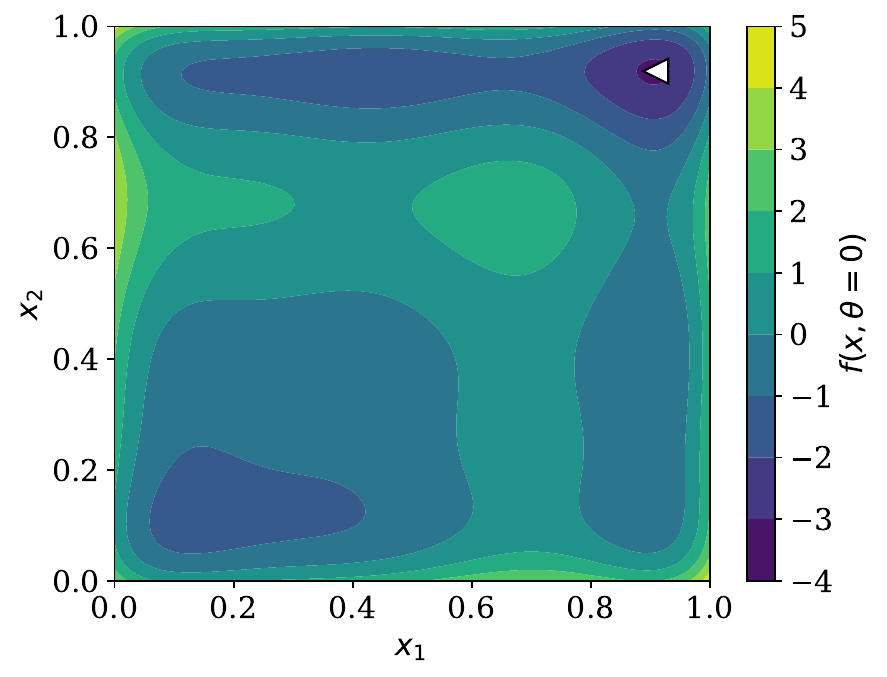}
        \caption{Synthetic Polynomial, \(\bm{\theta} = \begin{bmatrix}
            0 & 0
        \end{bmatrix}\)}%
        \label{fig:synth_poly1}
    \end{subfigure}
    \hfill
    \begin{subfigure}[t]{0.45\textwidth}
        \centering
        \includegraphics[width=\textwidth]{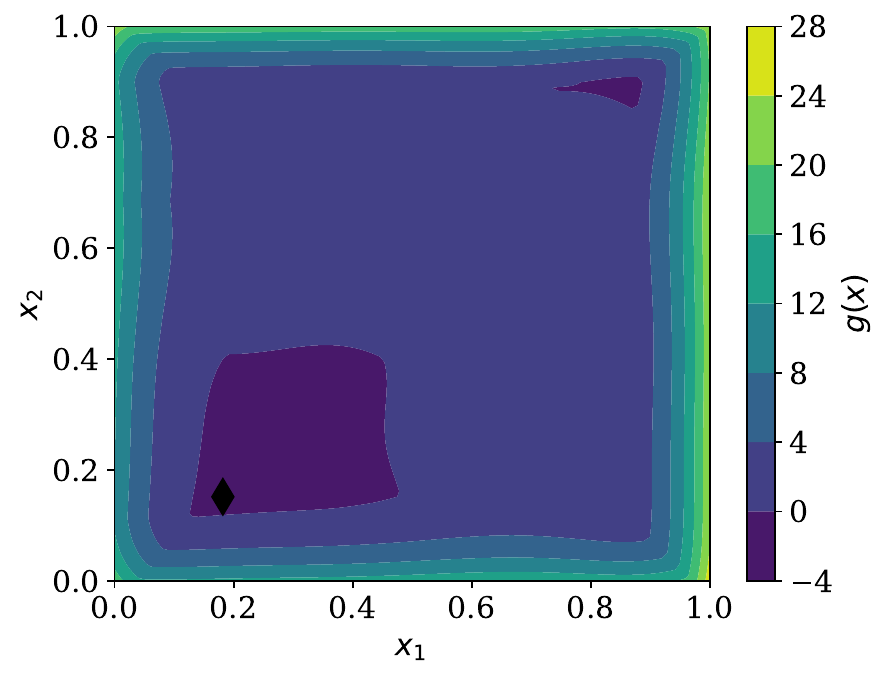}
        \caption{Robust Synthetic Polynomial}%
        \label{fig:synth_poly2}
    \end{subfigure}
    \caption{Visualization of the robust variant Synthetic Polynomial Problem. The global robust optimum is indicated by $\blacklozenge$, the global minimum by $\lhd$.}%
    \label{fig:synth_poly_p}
\end{figure}

Figure~\ref{fig:synth_poly1} shows the original optimization problem (\(\theta = 0\)). The maximizing function \(g(\bm{x}) = \max_{\bm{\theta}} f(\bm{x}, \bm{\theta})\) is visualized in figure~\ref{fig:synth_poly2}. The robust optimum is far from the non-disturbed one.

Similar to \cite{Bogunovic}, we fix the hyperparameters to values found by maximum likelihood estimation using 500 randomly sampled points with function values below \(15\). 
The model is initialized with ten random points from the domain.
We run each algorithm with 100 different initializations for 100 iterations.

\section{Additional Experiment Results}
\subsection{Runtime Results}
\subsubsection{Time Complexity of Algorithm}
\label{sec:theoretical_runtime}
To evaluate the time complexity of one iteration of \gls{RES}, we have to consider the types of parameters, i.e. wether the (un)controllable parameters are discrete or continuous.
Therefore, we distinguish four cases: the case of fully discrete parameters, the case of fully continuous parameters and the mixed ones.

For all of them, the calculation time is dominated by calculation of equation~(\ref{eq:prediction}). Two aspects are influencing it: the marginalization step, which is of order \(\mathcal{O}\left(N^3\right)\), with \(N\) being the number of data points in \(D_t\), and the optimization procedure to find the argmax function \(\bm{h}_c(\bm{x})\) and the corresponding function value \(g_c\left(\bm{x}\right)\), with a single prediction of the function sample \(f_c\) from a set of \(C\) function samples, each with $F$ Fourier feature functions scaling with \(\mathcal{O}\left(F^2\right)\).

An additional aspect to take into account is the scaling of the different applied optimization procedures. We apply the Nelder-Mead method, which (in the worst case of a nonconvex and nonsmooth function) scales with $\mathcal{O}\left(\frac{d^2}{\psi^4}\right)$ to reach a required precision $\psi$ in dimensionality $d$ \citep{complexity_nelder_mead}, the L-BFGS-B algorithm, which scales with maximum order $\mathcal{O}\left(d\right)$ per iteration \citep{Zhu1997}, and simple maximization of $N_d$ data points, being of $\mathcal{O}\left(N_d\right)$. Given these prerequisites, we can derive the complexity for all cases of parameter type combinations. In the following, we refer to the dimensionality of the uncontrollable parameters as \(d_u\), to the dimensionality of the controllable parameters as \(d_c\), to the number of uncontrollable parameters as \(N_u\), and to the number of controllable parameters as \(N_c\).

\paragraph{Fully continuous parameters.}
We search for the maximum of the acquisition function by multistart Nelder-Mead method with \(N_R\) restarts.
In each Nelder-Mead iteration, we have to call multistart the L-BFGS-B optimizer with \(N_r\) restarts and \(N_i\) iterations.
Therefore, we find a complexity of \(N_R \mathcal{O}\left(\frac{\left(d_c + d_u\right)^2}{\psi^4}\right) C \left(\mathcal{O}\left(N^3\right) +  N_r N_i \left(\mathcal{O}\left(d_u\right) + \mathcal{O}\left(F^2\right)\right)\right)\).

\paragraph{Fully discrete parameters.}
In the fully discrete case, we have to evaluate all combinations of parameters and maximize afterward. For each controllable parameter, we have to find the maximizing value of the uncontrollable parameters. Therefore, we have to predict once and maximize \(N_c\) times.
Therefore, we find a complexity of \(\mathcal{O}\left(N_c N_u\right) + C \left(\mathcal{O}\left(N^3\right) + \mathcal{O}\left(F^2\right) + N_c \mathcal{O}\left(N_u\right)\right)\).

\paragraph{Continuous controllable parameters and discrete uncontrollable parameters.}
In this case, we optimize the acquisition function again via multistart Nelder-Mead method but find the maximizing uncontrollable parameters in the discrete way. In each Nelder-Mead iteration, we have to maximize the function sample, and we find a complexity of \(N_R \mathcal{O}\left(\frac{d_c^2}{\psi^4}\right) C \left(\mathcal{O}\left(N^3\right) +  \mathcal{O}\left(F^2\right) + \mathcal{O}\left(N_u\right)\right)\).

\paragraph{Discrete controllable parameters and continuous controllable parameters.}
Even though we do not perform experiments for this case, we provide the result for sake of completeness. Here, the outer optimization is performed in a discrete manner, while the inner one is continuous, so the complexity scales with 
\(\mathcal{O}\left(N_c\right) + N_c \left(C \left(\mathcal{O}\left(N^3\right) + N_r N_i \left(\mathcal{O}\left(d_u\right) + \mathcal{O}\left(F^2\right)\right)\right)\right)\).

\subsubsection{Practical Runtime Experiments}
\label{sec:practical_runtime}
In tables~\ref{tab:runtime_within_model},~\ref{tab:runtime_hartmann3d}, and~\ref{tab:runtime_branin} we summarize the computation time of the algorithms for a fully continuous experiment (e.g., the within-model comparison), for a fully discretized experiment (e.g., the discretized Hartmann function), and an experiment that has a continuous space of controllable parameters \(\mathcal{X}\) and a discrete space of uncontrollable parameters \(\Theta\), (e.g., the Branin function).
The measured runtime contains the initialization and the optimization of the acquisition function for one iteration.
For \sopt{}, we include the runtime for all values of the exploration constant \(\sqrt{\beta}\). The experiments are taken out on Intel Xeon Gold 5118 CPUs, using 12 cores in parallel, for the Branin function, we were able to apply 24 cores.

Overall, the runtime of our approach \gls{RES} is between \sopt{} and \gls{KG}, with \gls{KG} being faster on the Branin function, which is due to it running on a small discrete space of controllable parameters \(\mathcal{X}\) and \gls{RES} on a continuous space of controllable parameters \(\mathcal{X}\). We assume that the optimization of the \gls{RES} acquisition function on the mixed space \(\mathcal{X} \times \Theta\) takes more iterations than the optimization over its discrete version.  
{\begin{table}
    \centering
    \caption{Runtime results for within model comparison. Results in seconds. $^\ast$: \gls{KG} algorithm runs on a discretized space of \(50 \times 50\).}\label{tab:runtime_within_model}
    \begin{tabular}{rlllllll}
      \toprule 
      \bfseries Quantile & \bfseries \gls{RES} & \bfseries \sopt{} & \bfseries \gls{MES} & \bfseries \gls{KG}$^\ast$ & \bfseries \gls{UCB} & \bfseries \gls{EI} \\
      \midrule 
      25~\% & 203.99 & 34.32 & 0.34 & 1197.04 & 0.12 & 0.17\\
      50~\% & 232.97 & 44.92 & 0.39 & 1525.47 & 0.15 & 0.22\\
      75~\% & 271.23 & 58.87 & 0.52 & 2043.63 & 0.24 & 0.39\\
      \bottomrule 
    \end{tabular}
\end{table}
\begin{table}
    \centering
    \caption{Runtime results for the fully discretized Hartmann function. Results in seconds.}\label{tab:runtime_hartmann3d}
    \begin{tabular}{rlllllll}
      \toprule 
      \bfseries Quantile & \bfseries \gls{RES} & \bfseries \sopt{} & \bfseries \gls{MES} & \bfseries \gls{KG} & \bfseries \gls{UCB} & \bfseries \gls{EI} \\
      \midrule 
      25~\% & 195.413 & 0.021 & 0.117 & 1199.908 & 0.021 & 0.022\\
      50~\% & 199.438 & 0.026 & 0.125 & 2278.777 & 0.026 & 0.026\\
      75~\% & 204.000 & 0.031 & 0.133 & 3848.871 & 0.031 & 0.031\\
      \bottomrule 
    \end{tabular}
\end{table}
\begin{table}[H]
    \centering
    \caption{Runtime results for the Branin function. Results in seconds. $^\ast$: \gls{KG} algorithm runs on a discretized space of \(50 \times 1\).}\label{tab:runtime_branin}
    \begin{tabular}{rlllllll}
      \toprule 
      \bfseries Quantile & \bfseries \gls{RES} & \bfseries \sopt{} & \bfseries \gls{MES} & \bfseries \gls{KG}$^\ast$ & \bfseries \gls{UCB} & \bfseries \gls{EI} \\
      \midrule 
      25~\% & 28.069 & 1.309 & 0.156 & 9.819 & 0.066 & 0.084\\
      50~\% & 33.536 & 3.593 & 0.167 & 12.373 & 0.072 & 0.093\\
      75~\% & 38.387 & 5.305 & 0.177 & 15.262 & 0.079 & 0.103\\
      \bottomrule 
    \end{tabular}
\end{table}}

\subsubsection{Performance over Runtime}
\label{sec:performance_runtime}
\begin{figure}[H]
    \centering
    \includegraphics[width=0.6\textwidth]{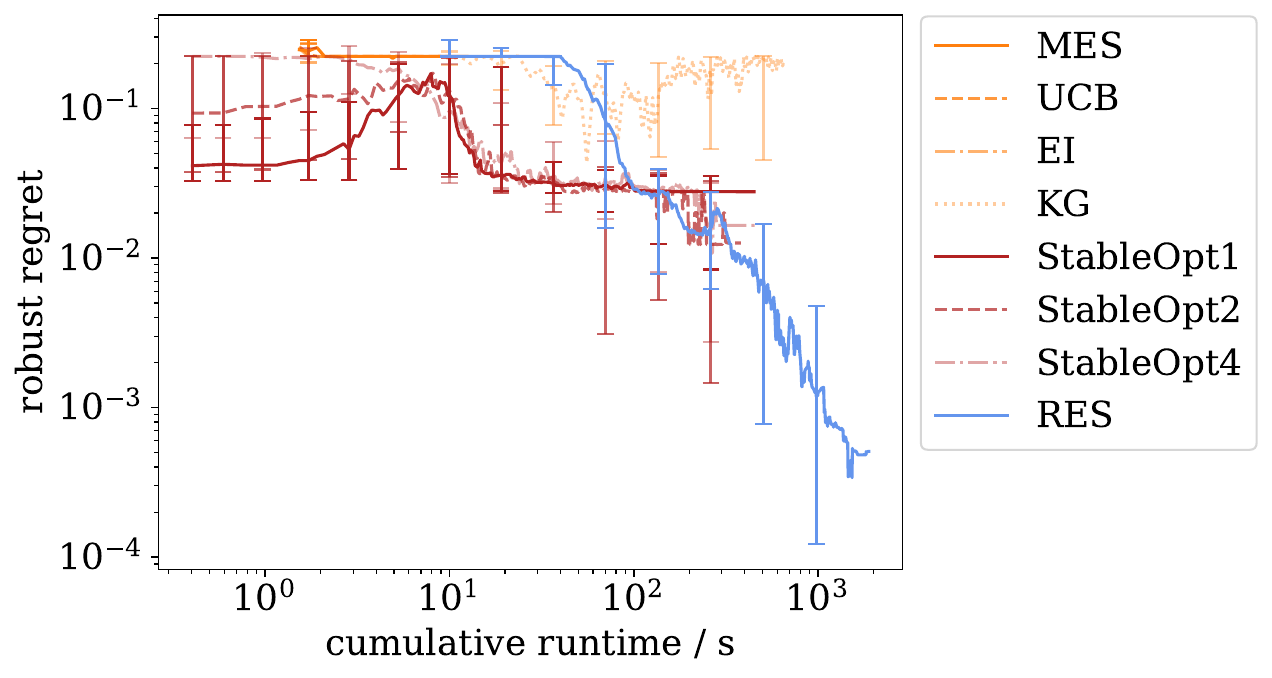}
    \caption{Regret over runtime for the Branin Problem. \gls{RES} reaches the same robust regret as \sopt{} in a similar amount of time.}
    \label{fig:regret_runtime}
\end{figure}

In figure~\ref{fig:regret_runtime}, we provide the robust regret over the runtime for the Branin function. Even though \gls{RES} experiences a slow start, it achieves a similar regret in the same time as \sopt{} with a significantly lower number of iterations (as can be seen from the experiment in the main part of the paper).

\subsection{Results for Robust Robot Pushing Problem}
\label{sec:robot_pushing}
We adopt the robust robot pushing problem from \cite{Bogunovic}, which is based on the publicly available code\footnote{\texttt{https://github.com/zi-w/Max-value-Entropy-Search}} of the robot pushing objective by \cite{Wang17}. 

In the problem, a good pre-image for pushing an object to an unknown target location is sought. Precisely, there are two different target locations, where the first is uniformly distributed over the domain and the second uniform over the \(l_1\)-ball centered at the first target location with radius \(r=2.0\).
Each evaluation calls a function \(f(r_x, r_y, r_t) = 5 - d_{end}\), where \((r_x, r_y) \in \left[-5, 5\right]^2\) is the initial robot location, \(r_t \in \left[1, 30\right]\) is the pushing duration and \(d_{end}\) is the distance to the target location. 

\begin{figure}
    \centering
    \includegraphics[width=0.6\textwidth]{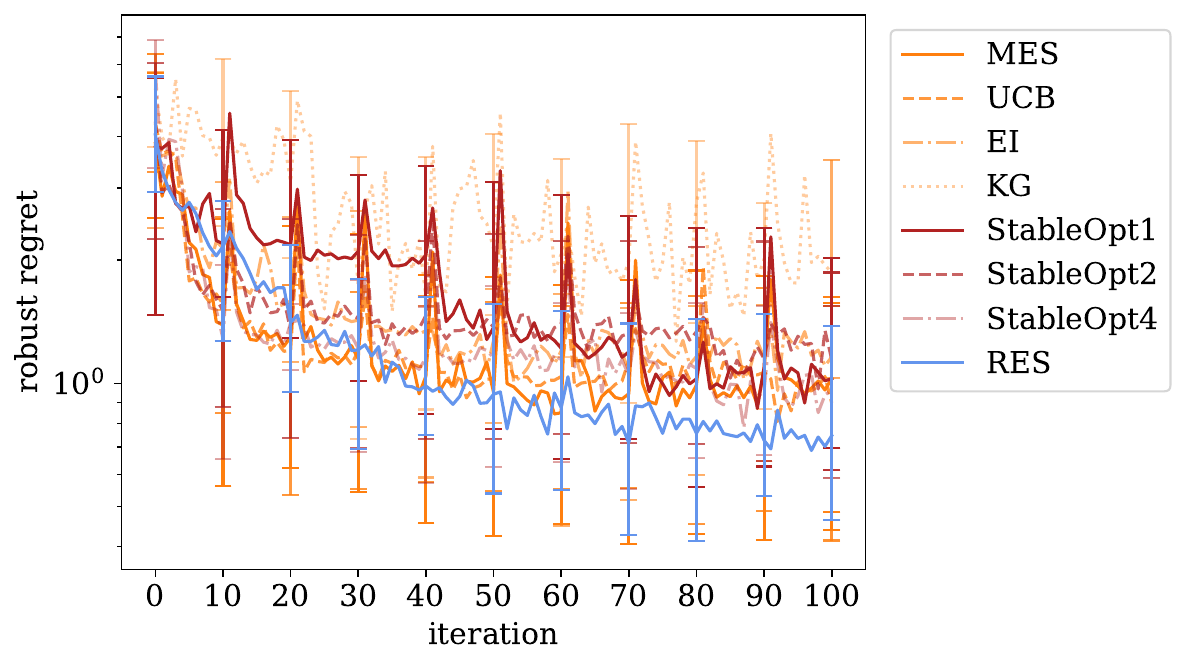}
    \caption{Results for robust robot pushing problem.}
    \label{fig:robot_pushing}
\end{figure}

We run the problem 30 times for 100 iterations, where each initialization consists of a randomly drawn pair of targets and two starting positions, one for each target. We make a fully Bayesian treatment of the model hyperparameters, updated every 10th iteration. In figure~\ref{fig:robot_pushing}, we report the robust regret: \gls{RES} again shows a superior performance. The large discontinuities in the curves are caused by hyperparameter re-estimation.

\end{document}